\PassOptionsToPackage{unicode}{hyperref}
\PassOptionsToPackage{hyphens}{url}
\PassOptionsToPackage{dvipsnames,svgnames,x11names}{xcolor}

\documentclass[
  11pt,
  a4paper,
  fontset=fandol
]{ctexart}

\ctexset{
    subparagraph = {
        format = \normalfont\normalsize\bfseries,
        runin = true,
        beforeskip = 0.8 ex,
        indent = 0pt,
        afterskip = 2\ccwd
    }
}

\ctexset{
    paragraph/format = \normalfont\normalsize\bfseries,
    subparagraph/format = \normalfont\normalsize\bfseries
}

\let\subsubsubsection\paragraph
\let\subsubsubsubsection\subparagraph

\newcounter{subsubparagraph}[subparagraph]
\renewcommand{\thesubsubparagraph}{%
    \roman{subsubparagraph})%
}
\newcommand{\subsubparagraph}[1]{%
    \par
    \addvspace{0.1ex}%
    \refstepcounter{subsubparagraph}%
    \noindent
    \textit{\thesubsubparagraph\ #1}%
    \hspace{1\ccwd}%
    \ignorespaces
}

\usepackage{xcolor}
\usepackage[
  top=2.2cm,
  bottom=2.2cm,
  left=2.4cm,
  right=2.4cm
]{geometry}

\usepackage{amsmath,amssymb}

\usepackage{iftex}

\ifPDFTeX
  \usepackage[T1]{fontenc}
  \usepackage[utf8]{inputenc}
  \usepackage{textcomp}
\else
  \usepackage{unicode-math}
  \defaultfontfeatures{Scale=MatchLowercase}
  \defaultfontfeatures[\rmfamily]{Ligatures=TeX,Scale=1}
\fi

\usepackage{lmodern}
\usepackage{fancyvrb}
\usepackage{listings}
\usepackage{changepage}

\ifPDFTeX\else
\fi
\IfFileExists{upquote.sty}{\usepackage{upquote}}{}
\IfFileExists{microtype.sty}{
  \usepackage[]{microtype}
  \UseMicrotypeSet[protrusion]{basicmath} 
}{}
\usepackage{setspace}
\makeatletter
\@ifundefined{KOMAClassName}{
  \IfFileExists{parskip.sty}{%
    \usepackage{parskip}
  }{
    \setlength{\parindent}{0pt}
    \setlength{\parskip}{6pt plus 2pt minus 1pt}}
}{
  \KOMAoptions{parskip=half}}
\makeatother
\usepackage{longtable,booktabs,array}
\usepackage{caption}

\usepackage{booktabs}
\usepackage{tabularx}
\usepackage[table]{xcolor}
\usepackage{multirow}
\usepackage{calc} 
\usepackage{etoolbox}
\AtBeginEnvironment{longtable}{\footnotesize}
\makeatletter
\patchcmd\longtable{\par}{\if@noskipsec\mbox{}\fi\par}{}{}
\makeatother
\IfFileExists{footnotehyper.sty}{\usepackage{footnotehyper}}{\usepackage{footnote}}
\makesavenoteenv{longtable}
\usepackage{graphicx}
\makeatletter
\newsavebox\pandoc@box
\newcommand*\pandocbounded[1]{
  \sbox\pandoc@box{#1}%
  \Gscale@div\@tempa{\textheight}{\dimexpr\ht\pandoc@box+\dp\pandoc@box\relax}%
  \Gscale@div\@tempb{\linewidth}{\wd\pandoc@box}%
  \ifdim\@tempb\p@<\@tempa\p@\let\@tempa\@tempb\fi
  \ifdim\@tempa\p@<\p@\scalebox{\@tempa}{\usebox\pandoc@box}%
  \else\usebox{\pandoc@box}%
  \fi%
}
\def\fps@figure{htbp}
\makeatother

\usepackage[normalem]{ulem}

\usepackage{bookmark}
\IfFileExists{xurl.sty}{\usepackage{xurl}}{} 
\makeatletter
\@ifundefined{xmpquote}{}{}
\makeatother
\hypersetup{
  pdftitle={SeetaPsych v1.0: An Open-source Computer Vision Toolkit for Behavior-based Psychological Measurement},
  colorlinks=true,
  linkcolor={blue},
  filecolor={Maroon},
  citecolor={Blue},
  urlcolor={blue},
  pdfcreator={LaTeX via pandoc}}

\title{SeetaPsych v1.0: An Open-source Computer Vision Toolkit for \\  Behavior-based Psychological Measurement}
\usepackage{etoolbox}
\makeatletter
\providecommand{\subtitle}[1]{
  \apptocmd{\@title}{\par {\large #1 \par}}{}{}
}
\makeatother
\author{
    Jiabei Zeng$^{1,*}$ \quad
    Chiqin Li$^{1,*}$ \quad
    Kaizhou Li$^{1,3,*}$ \quad
    Fei Chang$^{1,*}$ \quad 
    Yong Li$^{2,*}$ \quad  \\ 
    Yuanhao Zhao $^{1,*}$ \quad
    Dan Han$^{1,*}$ \quad
    Wenqiang Yang $^{1,*}$ \quad \\
    Xilin Chen$^{1,\dagger}$ \quad
    Shiguang Shan$^{1,3,\dagger}$
    \\[0.8em]
    \small $^{1}$Institute of Computing Technology, Chinese Academy of Science, Beijing, China
    \\
    \small $^{2}$ Southeast University, Nanjing, China
    \\
    \small $^{3}$ Beijing Seetatech Co., Ltd., Beijing, China
    \\[0.5em]
    \small $^{*}$Equal contribution.
    \qquad
    $^{\dagger}$Project leader.
}

\date{}

\begin{document}

\setlength{\parindent}{2\ccwd}
\makeatletter
\def\@afterindentfalse{\let\if@afterindent\iffalse}
\makeatother

\maketitle
\pagestyle{plain}

\begin{abstract}
Automated visual analysis opens new avenues for behavior-based psychological measurement. Nevertheless, existing technological modules are typically scattered across task-specific systems with heterogeneous interfaces and disparate deployment requirements. In this work, we present SeetaPsych v1.0, an open-source, unified and extensible computer vision toolkit designed to extract psychologically relevant signals from facial images and/or face-based videos. The current release encompasses four major core modules aiming at behavior-based physiological perception: unified face-based emotion analysis (simultaneous facial expression recognition, facial action unit detection, and valence–arousal estimation), camera-based heart-rate estimation, screen point-of-gaze estimation, and scene gaze following. A suite of auxiliary preprocessing modules for human centric visual analysis is also included, comprising face detection, facial landmark detection, and head detection. These functionalities are encapsulated within a modular Pipeline/Runner architecture that automatically resolves attribute dependencies, constructs computation graphs, and support intermediate-result sharing among modules. SeetaPsych provides standardized Python APIs to facilitate reproducible, large scale analyses, alongside an interactive WebUI for rapid, code-free method evaluation. Overall, SeetaPsych offers an integrated and accessible visual measurement platform for research in psychology, behavioral science, human–computer interaction, and related fields.  
\end{abstract}

\clearpage
\tableofcontents
\clearpage

\setstretch{1.05}
\section{Introduction}

Human faces and bodies provide rich signals that reveal human psychological states, physiological conditions, and observable behaviors. For example, facial expressions and subtle facial movements convey affective states and behavioral responses, while gaze provides important cues about visual attention, social engagement, and interaction intent. Physiological indicators, such as heart rate, are closely associated with arousal, stress, and autonomic nervous system activity. In addition, body movements and interpersonal gaze behaviors reveal patterns of engagement, coordination, and social interaction dynamics. These diverse signals have long been central to research in psychology, behavioral science, human-computer interaction, and related fields.

Despite recent advances in automated face and body signal analysis, computer vision techniques remain underutilized in psychological and behavioral research due to the lack of integrated and accessible measurement tools. Traditionally, such studies have relied on manual observation, expert annotation, questionnaires, or specialized sensing devices to measure human-related signals. Recent computer vision methods have enabled the automated extraction of psychologically relevant signals from images and videos. However, existing solutions are typically developed as independent systems for specific tasks, including facial expression recognition, gaze estimation, remote heart-rate measurement, and gaze-following analysis. These systems often differ in software interfaces, input requirements, and deployment procedures, making their integration into unified experimental pipelines technically challenging, especially for researchers without a computer science background.

To address this limitation, we introduce SeetaPsych v1.0, a unified and open-source toolkit for automated psychological measurement from visual data. We provide source code and pretrained models to facilitate reproducible research and community-driven development. The current version focuses on facial and face-related measurements, including face-based emotion analysis, sceen point-of-gaze estimation, and camera-based heart-rate estimation, and also supports person-scene visual attention analysis, such as gaze following (also known as gaze target detection). The toolkit is designed to be extensible, with planned support for additional body-related behavioral signals, including subtle body movements and other nonverbal cues.

To facilitate practical adoption in psychological experiments, SeetaPsych provides two complementary usage modes. First, an interactive graphical user interface enables researchers to apply the included visual analysis methods to experimental images or videos and visualize the resulting outputs, allowing rapid evaluation of their applicability to specific experimental settings without programming. Second, standardized application programming interfaces (APIs) support batch processing and large-scale automated analysis, facilitating integration into reproducible experimental and data-analysis pipelines.

Overall, SeetaPsych is designed as a human-centric visual perception toolkit for behavior-based psychological measurement, supporting applications in psychology, behavioral science, human-computer interaction, and related fields. By providing open-source implementations and models, SeetaPsych promotes reproducible and accessible vision-based psychological measurement. The following sections present an overview of the SeetaPsych architecture, describe its functional modules with algorithmic details and performance evaluations, and introduce the usage workflow, including installation, APIs, and WebUI.

\section{SeetaPsych Overview}

\begin{figure}[t]
\centering
\includegraphics[width=5.75913in,height=4.08861in]{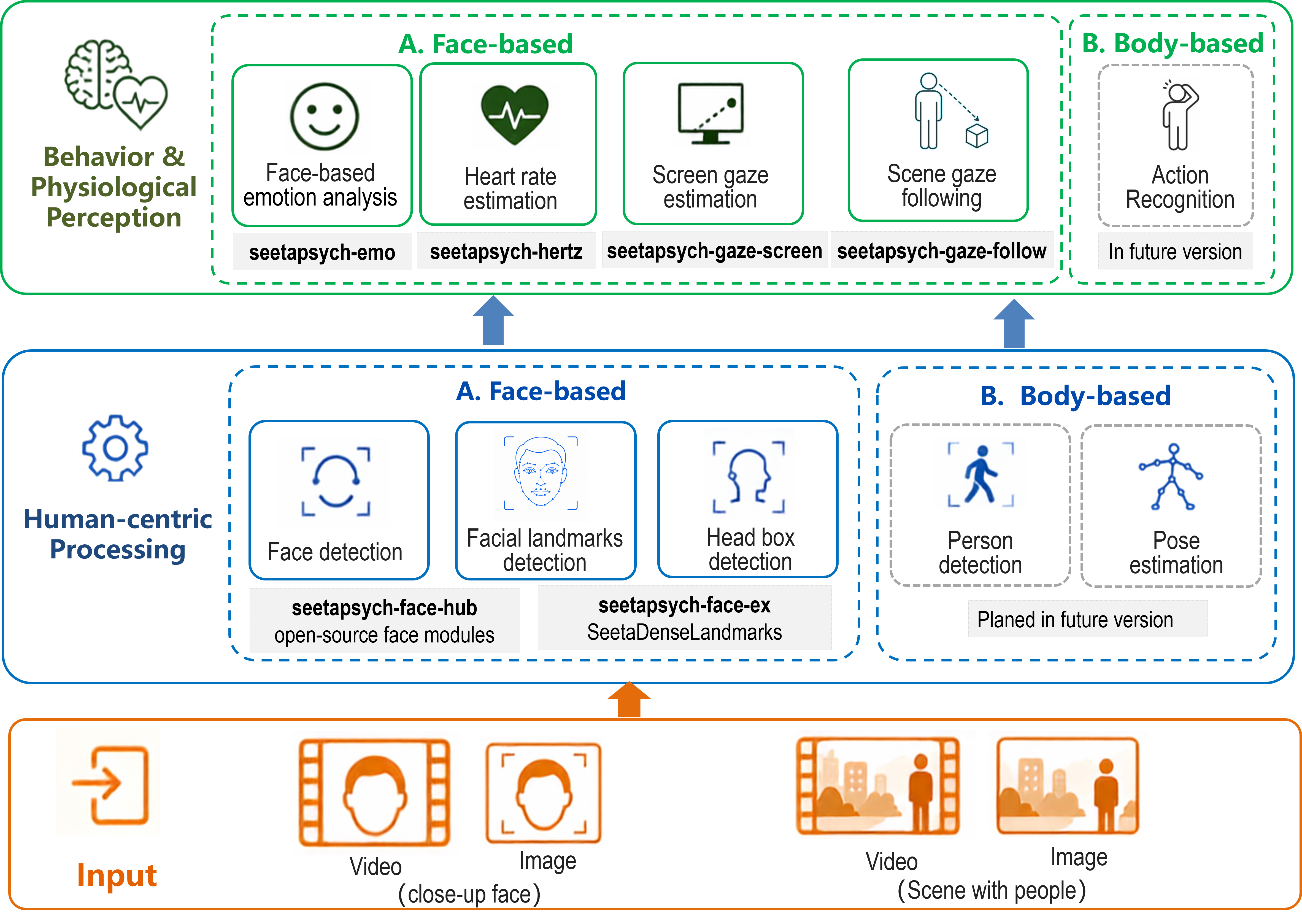}
\caption{Overview of the functional modules in SeetaPsych}
\label{fig:seetapsych}
\end{figure}

SeetaPsych is organized around two groups of visual analysis capabilities: Human-centric Processing and Behavior \& Physiological Perception, as illustrated in Figure~\ref{fig:seetapsych}. Human-centric Processing provides the preprocessing components required to localize and represent human-related visual structures, while Behavior\& Physiological Perception extracts higher-level behavioral and physiological signals from these intermediate representations.

In SeetaPsych v1.0, the Behavior \& Physiological Perception layer provides four major functions, with their corresponding algorithms organized into task-oriented sub-projects. 
Table~\ref{tab:seetapsych_modules} summarizes the functions, algorithms/models, and corresponding sub-projects included in SeetaPsych v1.0.
Face-based emotion analysis jointly estimates facial expressions, facial action units, and valence–arousal values, and is provided in sub-project \texttt{seetapsych-emo}, which integrates the SeetaEmoNetwork. Heart-rate estimation extracts physiological signals from facial videos and is provided in \texttt{seetapsych-herts}, which includes the AdaChrom and TinyHR methods. Screen point-of-gaze estimation predicts a user's gaze location on a screen and is provided in \texttt{seetapsych-gaze-screen}. The current release supports direct 2D point-of-gaze estimation using the open-source methods AFFNet\cite{bao2021affnet}, as well as 3D gaze-based estimation using TdGazeNet. 
Scene gaze following, provided in \texttt{seetapsych-gaze-follow}, estimates where a person is looking in a scene using CoSI-Gaze. In addition to these higher-level functions, SeetaPsych provides supporting human-centric preprocessing components. 
\texttt{Seetapsych-face-hub} integrates commonly used open-source face-processing methods, including RetinaFace\cite{deng2020retinaface} (InsigeFace version~\cite{insightface} and a PyTorch implementation~\cite{pytorch_retinaface}), ArcFace~\cite{deng2019arcface,insightface}, and MediaPipe Face Mesh~\cite{kartynnik2019facemesh,mediapipe}, while \texttt{seetapsych-face-ex} provides the internally developed SeetaDenseLandmarks model for 280-point dense facial landmark detection. Body-based behavioral analysis is planned for future versions. The detailed architectures, training procedures, and evaluations of these algorithms are presented in Section~\ref{sec:func_modules}.

\begin{table}[htbp]
    \centering
    \caption{Summary of the functions, algorithms/models, and corresponding sub-projects included in SeetaPsych v1.0.}
    \label{tab:seetapsych_modules}

    \small
    \setlength{\tabcolsep}{4pt}
    \renewcommand{\arraystretch}{1.2}

    \begin{tabularx}{\linewidth}{
        >{\centering\arraybackslash}m{0.24\linewidth}
        >{\centering\arraybackslash}m{0.46\linewidth}
        >{\centering\arraybackslash}m{0.22\linewidth}
    }
        \toprule
        \textbf{Function}
        & \textbf{Algorithms/Models}
        & \textbf{Sub-project} \\
        \midrule

        face-based emotion analysis
        & SeetaEmoNetwork
        & seetapsych-emo \\

        \rowcolor{gray!10}
        heart-rate estimation
        & AdaChrom, TinyHR
        & seetapsych-herts \\

        screen point-of-gaze estimation
        & AFFNet, TdGazeNet
        & seetapsych-gaze-screen \\

        \rowcolor{gray!10}
        scene gaze following
        & CoSI-Gaze
        & seetapsych-gaze-follow \\

        \multirow{2}{=}{\cellcolor{white}\centering preprocessing}
        & RetinaFace, ArcFace, MediaPipe Face Mesh, PyTorch RetinaFace
        & seetapsych-face-hub \\

        \rowcolor{gray!10}
        \cellcolor{white}
        & SeetaDenseLandmarks
        & seetapsych-face-ex \\

        \bottomrule
    \end{tabularx}
\end{table}

At the framework level, these components are exposed through a common attribute-based interface. Users specify the attributes to be obtained, and SeetaPsych automatically identifies the required algorithm modules, resolves their dependencies, and constructs the corresponding computation graph. The resulting pipeline is executed through the Runner, with intermediate results shared among dependent modules. Pipelines can be accessed programmatically through the Python API or configured interactively through the WebUI. The configuration, dependency resolution, installation, and execution workflow is described in Section~\ref{sec:usage}.

\section{Functional Modules}
\label{sec:func_modules}

The Behavior \& Physiological Perception modules cover both face-based and body-based analysis. In SeetaPsych V1.0, we provide four functional modules: face-based emotion analysis, heart rate estimation, screen point-of-gaze estimation, and scene gaze following. The following sections describe these four modules in detail.

\subsection{Face-based emotion analysis}

Facial expressions and subtle facial movements provide important cues for understanding emotional states and behavioral responses. To support comprehensive face-based emotion analysis, SeetaPsych v1.0 provides a unified module that jointly covers categorical expression recognition, facial action unit detection, and continuous valence–arousal estimation. This section introduces the supported outputs, the SeetaEmoNetwork model, and its evaluation on commonly used benchmark datasets.

\subsubsection{Functional Description}

The face-based emotion analysis module jointly performs facial expression recognition, facial action unit (AU) detection, and valence–arousal (VA) estimation from a facial image.

\subsubsection{Input and Output}

\textbf{Input:}
\begin{adjustwidth}{4em}{0em}
\textbf{Facial image:}
A cropped and aligned RGB facial image of size
\(
256 \times 256 \times 3
\).
Face cropping and alignment are performed using the preprocessing modules provided by SeetaPsych.
\end{adjustwidth}

\textbf{Outputs:}
{
\setlength{\leftmargini}{5em}
\begin{enumerate}
    \renewcommand{\labelenumi}{\arabic{enumi}.}
    \setlength{\itemsep}{0.5em}
    \setlength{\topsep}{0.4em}

    \item \textbf{Facial expression:}
    Probabilities for seven basic expression categories in the following order:
    neutral, anger, disgust, fear, happy, sad, and surprise.

    \item \textbf{Facial action units:}
    Occurrence probabilities for 16 action units (AUs) in the following order:
    AU1, AU2, AU4, AU5, AU6, AU7, AU9, AU10, AU12, AU15,
    AU17, AU20, AU23, AU24, AU25, and AU26. According to well-known Facial Action Coding System (FACS) \cite{ekman1978facial}, the descriptions of AUs are shown in Table~\ref{tab:description_of_aus}.

\begin{table}[htbp]
    \centering
    \caption{Descriptions of the 16 output AUs.}
    \label{tab:description_of_aus}
    \begin{tabular}{ll}
        \toprule
        AU & Description \\
        \midrule
        AU1  & Inner Brow Raiser \\
        AU2  & Outer Brow Raiser \\
        AU4  & Brow Lowerer \\
        AU5  & Upper Lid Raiser \\
        AU6  & Cheek Raiser \\
        AU7  & Lid Tightener \\
        AU9  & Nose Wrinkler \\
        AU10 & Upper Lip Raiser \\
        AU12 & Lip Corner Puller \\
        AU15 & Lip Corner Depressor \\
        AU17 & Chin Raiser \\
        AU20 & Lip Stretcher \\
        AU23 & Lip Tightener \\
        AU24 & Lip Pressor \\
        AU25 & Lips Part \\
        AU26 & Jaw Drop \\
        \bottomrule
    \end{tabular}
\end{table}

    \item \textbf{Valence and arousal:}
    Two continuous values representing valence and arousal.

\end{enumerate}
}

\subsubsection{Core Algorithms and Models}

\begin{figure}[t]
    \centering
    \includegraphics[width=5.15022in,height=2.91442in]{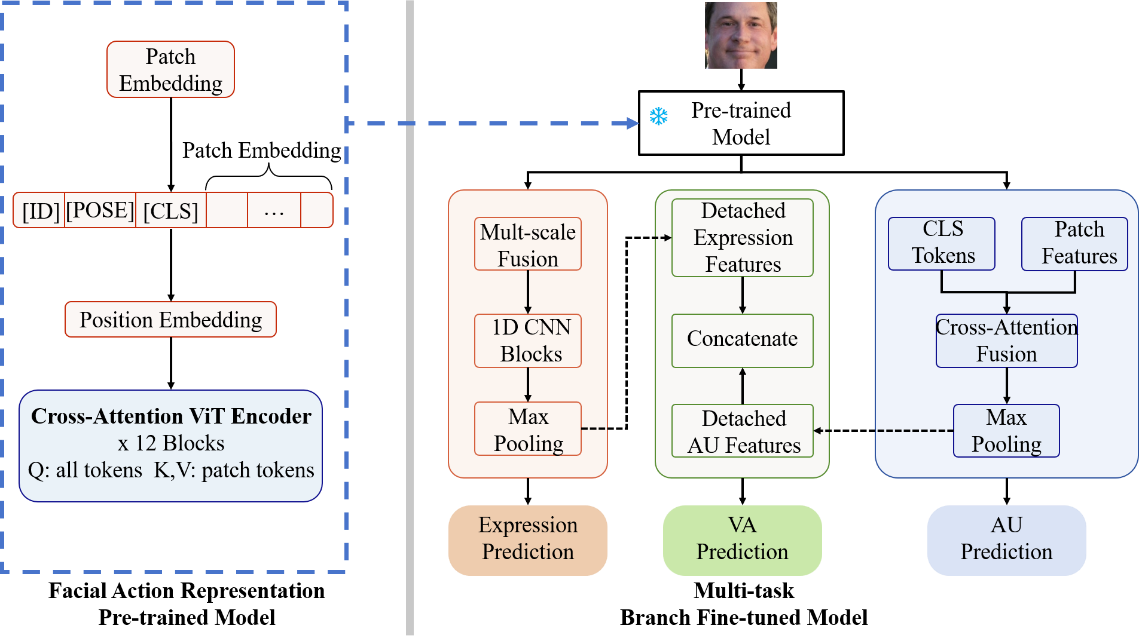}
    \caption{Overall framework of SeetaEmoNetwork.}
    \label{fig:seetaemo}
\end{figure}

SeetaPsych V1.0 employs SeetaEmoNetwork for face-based emotion analysis. The overall framework is illustrated in Figure~\ref{fig:seetaemo}, where the model inference pipeline corresponds to the multi-task branch fine-tuning stage. Facial action representations are first extracted using a pretrained facial action representation model and subsequently fed into the facial expression recognition and action unit (AU) detection branches to generate task-specific predictions. As for valence-arousal (VA) estimation, facial expression and AU features are jointly leveraged to predict valence and arousal.

Specifically, for the facial expression recognition branch, the complementary semantic information encoded at different feature levels is exploited. A multi-level fusion module is employed to selectively aggregate the outputs of cross-attention modules across multiple levels, followed by a 1D convolutional module to enhance local feature modeling. For the AU detection branch, global facial action information is incorporated to strengthen the representation of local facial movements. Specifically, patch embeddings from four layers are used as queries, while the CLS tokens from all layers are used as keys and values in the cross-attention operation. For the VA estimation branch, facial expressions provide coarse-grained localization in the VA space, whereas AUs characterize fine-grained facial muscle movements and facilitate the discrimination of subtle affective variations. Learning the mapping from facial expression and AU features to VA values is therefore more tractable than directly learning the mapping from high-dimensional visual representations to VA values. Accordingly, the VA estimation branch takes facial expression and AU features as input and predicts the corresponding valence and arousal values.

The model is trained in two stages: facial action representation pretraining and multi-task branch fine-tuning. The facial action representation pretraining stage consists of three steps. (1) Self-supervised feature initialization. A Masked Autoencoder (MAE) \cite{he2022mae} augmented with multiple learnable tokens is adopted and trained on AffectNet \cite{mollahosseini2017affectnet} to encourage the encoder to capture associations between facial regions and facial actions. (2) Multi-task supervised fine-tuning. The FEC dataset \cite{vemulapalli2019fec} is utilized to learn discriminative facial action representations while suppressing nuisance factors such as identity and head pose. (3) Self-supervised cross-reconstruction. The CelebV-Text dataset \cite{yu2023celebvtext} is used to further refine the encoder's ability to represent facial actions. After pretraining, the model produces ID tokens, CLS tokens, pose tokens, and patch embeddings.

During multi-task branch fine-tuning, the pretrained facial action representation model is frozen and used as a feature extractor. The extracted representations are then fed into three task-specific branches for expression recognition, action unit (AU) detection, and valence--arousal (VA) estimation. To fully exploit existing datasets, multiple datasets are integrated to construct the training data for each task. As illustrated in Fig. 1, the expression recognition and AU detection branches are trained first, followed by the VA estimation branch. To prevent VA training from affecting the first two branches, gradients from the VA branch are stopped at the expression and AU features. Specifically, AffectNet and RAF-DB \cite{li2017rafdb} are combined for expression recognition; EB+ \cite{ertugrul2019crossdomain}, GFT \cite{girard2017gft}, and DISFA \cite{mavadati2013disfa} are combined for AU detection; and the VA subset of AffectNet is used for VA estimation. The final trained model contains approximately 102M parameters.

\subsubsection{Performance}

To evaluate the performance of the proposed module, widely adopted benchmark datasets were employed. For the expression recognition task, the validation set of AffectNet and the test set of RAF-DB were utilized. For the AU detection task, three datasets, namely EB+, DISFA, and RAF-AU \cite{yan2020rafau}, were adopted. For the VA estimation task, the VA test set of AffectNet was used. Based on these datasets, extensive experiments were conducted.

\textbf{Expression Recognition Task:}The proposed method was compared with two open-source toolkits of the same category, Py-Feat \cite{cheong2023pyfeat} and LibreFace \cite{chang2024libreface}. The experimental results are presented in Table~\ref{tab:comparison_facial_expression}.

\begin{table}[htbp]
    \centering
    \caption{The overall accuracy and F1-scores compared with other methods for the expression recognition task.}
    \label{tab:comparison_facial_expression}
    \begin{tabular}{lcccc}
        \toprule
        \multirow{2}{*}{Method} 
        & \multicolumn{2}{c}{RAF-DB} 
        & \multicolumn{2}{c}{AffectNet} \\
        \cmidrule(lr){2-3} \cmidrule(lr){4-5}
        & Overall Acc. & F1 Score 
        & Overall Acc. & F1 Score \\
        \midrule
        Py-Feat   & --    & --   & --                 & 0.55 \\
        LibreFace & 0.83 & --   & 0.50 (8 classes)  & --   \\
        SeetaEmoNetwork  & 0.86 & 0.77 & 0.60 (7 classes)  & 0.60 \\
        \bottomrule
    \end{tabular}
\end{table}

The category-wise F1-score comparison results on the AffectNet dataset are presented in Table~\ref{tab:category_f1_affectnet}. Since LibreFace does not report class-specific experimental results, it was excluded from this comparison. As shown in Tables 2 and 3, compared with the two existing facial expression analysis toolkits, SeetaEmoNetwork achieves superior performance on both the AffectNet and RAF-DB datasets, demonstrating the competitiveness of the proposed method.

\begin{table}[htbp]
    \centering
    \caption{The category-wise F1-scores compared with Py-Feat on the AffectNet dataset.}
    \label{tab:category_f1_affectnet}
    \begin{tabular}{lccccccc}
        \toprule
        Method & Anger & Disgust & Fear & Happy & Sad & Surprise & Neutral \\
        \midrule
        Py-Feat  & 0.53 & 0.53 & 0.48 & 0.77 & 0.54 & 0.55 & 0.49 \\
        SeetaEmoNetwork & 0.62 & 0.50 & 0.59 & 0.78 & 0.60 & 0.53 & 0.55 \\
        \bottomrule
    \end{tabular}
\end{table}

The category-wise F1-score results on the RAF-DB dataset are presented in \ref{tab:category_f1_rafdb}. As shown in \ref{tab:category_f1_rafdb}, the F1-scores of SeetaEmoNetwork are below 0.7 only for the Disgust and Fear categories. This is mainly attributed to the limited number of samples for these two expressions in existing datasets, where the imbalanced data distribution poses challenges for accurate recognition. Nevertheless, SeetaEmoNetwork achieves satisfactory performance across the remaining expression categories.

\begin{table}[htbp]
    \centering
    \small
    \caption{The category-wise F1-scores on the RAF-DB dataset.}
    \label{tab:category_f1_rafdb}
    \begin{tabular}{lccccccc}
            \toprule
            Method & Anger & Disgust & Fear & Happy & Sad & Surprise & Neutral \\
            \midrule
            SeetaEmoNetwork & 0.70 & 0.56 & 0.63 & 0.95 & 0.86 & 0.85 & 0.86 \\
            \bottomrule
    \end{tabular}%
\end{table}

\textbf{AU Detection Task}. In this work, commonly used AUs that are beneficial for psychological state analysis were selected, resulting in the final adoption of the following 16 AUs: AU1, AU2, AU4, AU5, AU6, AU7, AU9, AU10, AU12, AU15, AU17, AU20, AU23, AU24, AU25, and AU26. Since this AU selection differs from those adopted in existing methods, direct comparisons with previous approaches were not conducted. Instead, the F1-scores of SeetaEmoNetwork on the EB+, DISFA, and RAF-AU datasets were reported, with the experimental results presented in Table~\ref{tab:F1_per_AU}. For the AU evaluation protocol, 20 subjects were randomly selected from the EB+ dataset to form the test subset. The DISFA dataset was partitioned according to the three-fold protocol used in previous work \cite{hu2026coin}, with the second fold adopted as the test set. Due to the limited number of samples in RAF-AU, the entire dataset was used for testing.

\begin{table}[htbp]
    \centering
    \caption{F1-score for each AU by SeetaEmoNetwork}
    \label{tab:F1_per_AU}
    \begin{tabular}{lccccccccc}
        \toprule
         & AU1 & AU2 & AU4 & AU5 & AU6 & AU7 & AU9 & AU10 & \\
        \midrule
        DISFA  & 0.51 & 0.55 & 0.75 & 0.56 & 0.53 & --   & 0.41 & --   & \\
        RAF-AU & 0.67 & 0.70 & 0.79 & 0.60 & 0.44 & 0.23 & 0.73 & 0.48 & \\
        EB+    & 0.32 & 0.38 & 0.70 & 0.40 & 0.84 & 0.84 & 0.61 & 0.87 & \\
        \midrule
         & AU12 & AU15 & AU17 & AU20 & AU23 & AU24 & AU25 & AU26 & Avg. \\
        \midrule
        DISFA  & 0.83 & 0.51 & 0.62 & 0.14 & --   & --   & 0.89 & 0.76 & 0.59 \\
        RAF-AU & 0.65 & 0.44 & 0.59 & 0.36 & 0.10 & 0.33 & 0.94 & 0.57 & 0.54 \\
        EB+    & 0.91 & 0.59 & 0.73 & 0.31 & 0.53 & 0.27 & --   & --   & 0.59 \\
        \bottomrule
    \end{tabular}
\end{table}

\textbf{VA Estimation Task}. The Concordance Correlation Coefficient (CCC) results on the VA subset of AffectNet datset are presented in Table~\ref{tab:affect_va}. Three existing methods, COIN \cite{hu2026coin}, MT-EmotiEffNet \cite{savchenko2023mtemotieffnet}, and VA-StarGAN \cite{kollias2020vastargan}, were selected for comparison. The results demonstrate that SeetaEmoNetwork outperforms all compared methods except COIN. COIN improves its performance by incorporating semantic information from CLIP \cite{radford2021clip}, whereas SeetaEmoNetwork achieves competitive results compared with state-of-the-art methods using only visual information.

\begin{table}[htbp]
    \centering
    \caption{The CCC results compared with other methods on the VA subset of the AffectNet dataset.}
    \label{tab:affect_va}
    \begin{tabular}{lcccc}
        \toprule
        Method & VA-StarGan & MT-EmotiEffNet & COIN & SeetaEmoNetwork \\
        \midrule
        Affect-VA & 0.55 & 0.57 & 0.60 & 0.59 \\
        \bottomrule
    \end{tabular}
\end{table}

Furthermore, the inference efficiency of SeetaEmoNetwork was evaluated. The model achieves an average inference speed of 4.6 ms per image on an NVIDIA RTX 3090 GPU.

\subsection{ Heart rate estimation}

Heart rate is an important physiological signal associated with arousal, stress, and autonomic activity, and can be estimated remotely from facial videos. SeetaPsych v1.0 provides contactless heart-rate estimation using two complementary approaches: the unsupervised signal-processing method AdaChrom and the learning-based model TinyHR. This section describes their input and output, algorithmic details, and evaluation on the VIPL-HR dataset.

\subsubsection{Functional Description}

The heart-rate estimation module estimates heart rate from facial video using remote photoplethysmography (rPPG). The current implementation supports per-frame heart-rate output.

\subsubsection{Input and Output}

\textbf{Input:}
\begin{adjustwidth}{4em}{0em}
\textbf{Facial video:}
A sequence of cropped RGB facial frames.
A frame rate of approximately 30 fps is recommended.
\end{adjustwidth}

\textbf{Output:}
\begin{adjustwidth}{4em}{0em}
\textbf{Heart rate:}
Per-frame heart-rate estimates expressed in beats per minute (BPM).
\end{adjustwidth}

\subsubsection{Core Algorithms and Models}

The heart-rate estimation module includes two implementations with distinct estimation pipelines, which are described in the following subsections.

\subsubsubsection{AdaChrom} 

\begin{figure}
\centering
\includegraphics[width=0.95\linewidth]{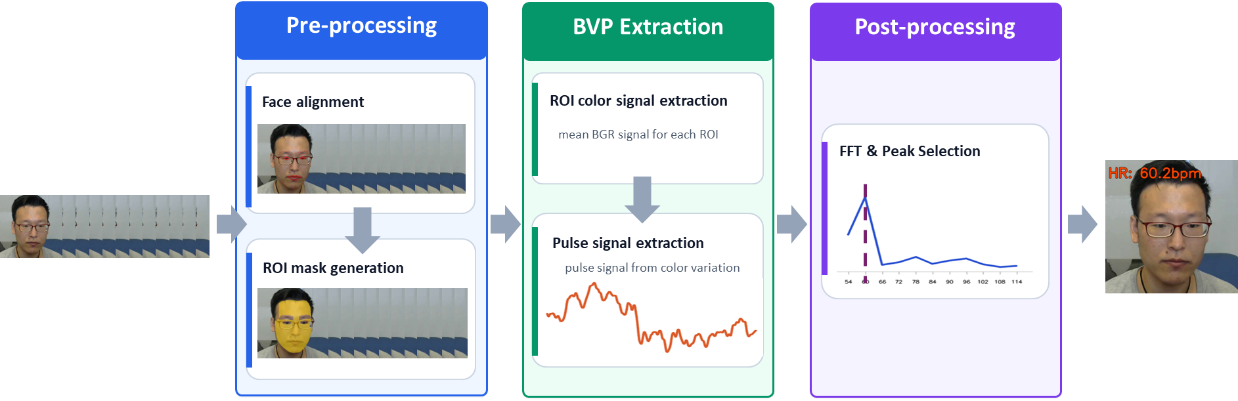}
\caption{Overview of the AdaChrom pipeline}
\label{fig:adachrom_pipeline}
\end{figure}

AdaChrom is an unsupervised remote photoplethysmography method for heart-rate estimation from facial videos. Instead of relying on labeled training data, AdaChrom estimates pulse-related blood volume pulse (BVP) signals from subtle temporal color variations in facial skin regions, providing an interpretable solution for contactless heart-rate estimation. 
Figure~\ref{fig:adachrom_pipeline} illustrates the overall pipeline of AdaChrom. 
Given a facial video sequence, AdaChrom follows a three-stage pipeline. 
In the pre-processing stage, face alignment is performed and ROI masks are generated. 
In the BVP extraction stage, mean BGR color signals are extracted from valid facial regions over a sliding temporal window, and the BVP signal is recovered from these temporal color signals. 
In the post-processing stage, heart rate is estimated from the recovered BVP signal through frequency-domain analysis and peak selection. 

\subsubsubsubsection{Pre-processing} 
In the pre-processing stage, AdaChrom establishes reliable spatial support for ROI-based signal extraction. Specifically, facial landmarks are aligned to reduce motion-induced ROI shifts, while valid facial skin regions are defined through ROI mask generation for subsequent BVP extraction.

\subsubparagraph{Face Alignment}
Face Alignment provides the geometric basis for all subsequent ROI operations. The aligned landmarks also provide a validity check for each frame, and frames with missing, zero, or non-finite landmark coordinates are excluded from the current estimation window. The details of face alignment can be found in SeetaPsych Face Hub.

\subsubparagraph{ROI Mask Generation}
After alignment, landmarks-defined polygon regions are used to construct facial ROI masks that isolate skin-dominant facial areas. A skin-color constraint is further applied in YCrCb color space. The final ROI mask is obtained by intersecting the geometric face mask with the detected skin mask. For each valid ROI, the algorithm records both the mean BGR value and the number of effective pixels, so empty or unreliable ROIs can be rejected before signal estimation. Four ROI strategies are supported in this implementation.

\textbf{Legacy YCrCb Skin ROI (AdaChrom-v1):} The legacy skin ROI first constructs a full-face mask from the facial landmarks and then applies a fixed YCrCb skin-color rule to identify skin pixels within the face region.

\textbf{Fixed Forehead-seeded Skin ROI (AdaChrom-v2):} The fixed forehead strategy uses a predefined landmark-based forehead region as the seed area for skin-color modeling. A two-dimensional Gaussian model is fitted in the Cr/Cb color space using the seed pixels, and pixels within the face region are retained when their color distribution is sufficiently close to the learned forehead skin model.

\textbf{Adaptive Forehead-seeded Skin ROI (AdaChrom-v3):} The adaptive forehead strategy extends the seed region around the forehead before estimating the Cr/Cb skin-color model. Compared with the fixed forehead seed, this strategy is designed to include more reliable forehead skin pixels and improve robustness when the fixed seed area is too small or partially affected by local landmark variation.

\textbf{Connected-component-filtered Skin ROI (AdaChrom-v4):} In addition to the adaptive forehead strategy, the connected-component strategy further filters the candidate skin regions according to spatial connectivity. Isolated false-positive regions are removed, while larger connected components that are spatially consistent with the face skin area are preserved for subsequent BGR signal extraction.

\subsubsubsubsection{BVP Extraction}
In the BVP extraction stage, AdaChrom converts spatial color observations from facial ROIs into temporal color traces and recovers the pulse-related BVP signal from these traces. This stage aggregates valid ROI measurements over a sliding temporal window, providing the signal representation used for subsequent heart-rate estimation.

\subsubparagraph{ROI Color Signal Extraction}

For every valid frame and ROI, the algorithm computes the spatial mean of the BGR pixel values inside the ROI mask~\cite{dehaan2013chrom}:

\[c_{t} = \ \left\lbrack {\overline{B}}_{t},\ {\overline{G}}_{t},\ {\overline{R}}_{t} \right\rbrack\]

This produces a temporal color trace for each ROI. Because remote photoplethysmography relies on subtle skin-color variations caused by blood volume changes, spatial averaging is used to suppress pixel-level noise and preserve the dominant temporal variation.

Before pulse extraction, the irregularly sampled color observations are regularized onto an evenly spaced time grid using the frame timestamps. The signal is then smoothed to reduce high-frequency noise and normalized by each channel's temporal mean to reduce illumination-scale effects.

\subsubparagraph{Pulse Signal Extraction}
The core BVP extraction model follows a CHROM-style chrominance projection~\cite{dehaan2013chrom}. The normalized RGB traces are transformed into two chrominance components:
\[X = 3R - 2G\]
\[Y = 1.5R + G - 1.5B\]

The pulse signal is then computed by balancing the two components according to their temporal standard deviations:
\[\alpha = \frac{std(X)}{std(Y)}\]
\[s(t) = X - \alpha Y\]

This projection emphasizes color changes associated with blood volume pulse while suppressing illumination variation and motion-related intensity changes. The projected BVP signal is then mean-centered, scaled, and smoothed before spectral analysis..

\subsubsubsubsection{Post-processing}
The post-processing stage converts the extracted BVP signal into the final heart-rate estimate through FFT-based spectral analysis and peak selection. A Hamming window is applied before the real-valued FFT to reduce spectral leakage. The magnitude spectrum is searched within the estimator's valid heart-rate range, approximately 50--120 bpm. The dominant spectral peak is selected as the heart-rate frequency and converted to beats per minute.
\[
f_{\mathrm{peak}} = \arg\max_{f \in \mathcal{F}_{HR}} \left| \mathrm{FFT}(s(t))(f) \right|
\]
\[
HR = 60 f_{\mathrm{peak}}
\]
where $s(t)$ denotes the extracted BVP signal, $f$ denotes the frequency variable in Hz, and $\mathcal{F}_{HR}$ denotes the valid heart-rate frequency range corresponding to approximately 50--120 bpm.

\subsubsubsection{TinyHR}
TinyHR is an end-to-end facial video model for estimating remote photoplethysmography (rPPG) signals and subsequently deriving heart rate from the predicted waveform. 

\subsubsubsubsection{Architecture}
The architecture is organized as a lightweight convolutional pipeline that progressively transforms frame-difference information into spatial-temporal physiological representations. The model consists of four main components: (1) a Frame Difference Fusion Stem, (2) Spatial Patch Embedding, (3) a Multi-scale Temporal Feature (MTF) Block, and (4) a waveform Predictor Head. During inference, the predicted rPPG waveform is further processed using deterministic band-pass filtering and spectral analysis to obtain the final heart-rate estimate.

The input is a facial video clip containing 160 RGB frames at a spatial resolution of $128\times 128$. The model first emphasizes short-term temporal color variations between adjacent frames, then extracts compact spatial features for each frame. These features are aggregated spatially into a temporal sequence and processed by multi-scale temporal convolutional operations, allowing the network to capture physiological variations over different temporal contexts while retaining the original frame rate.

\begin{figure}[htbp]
    \centering
    \includegraphics[width=0.95\linewidth]{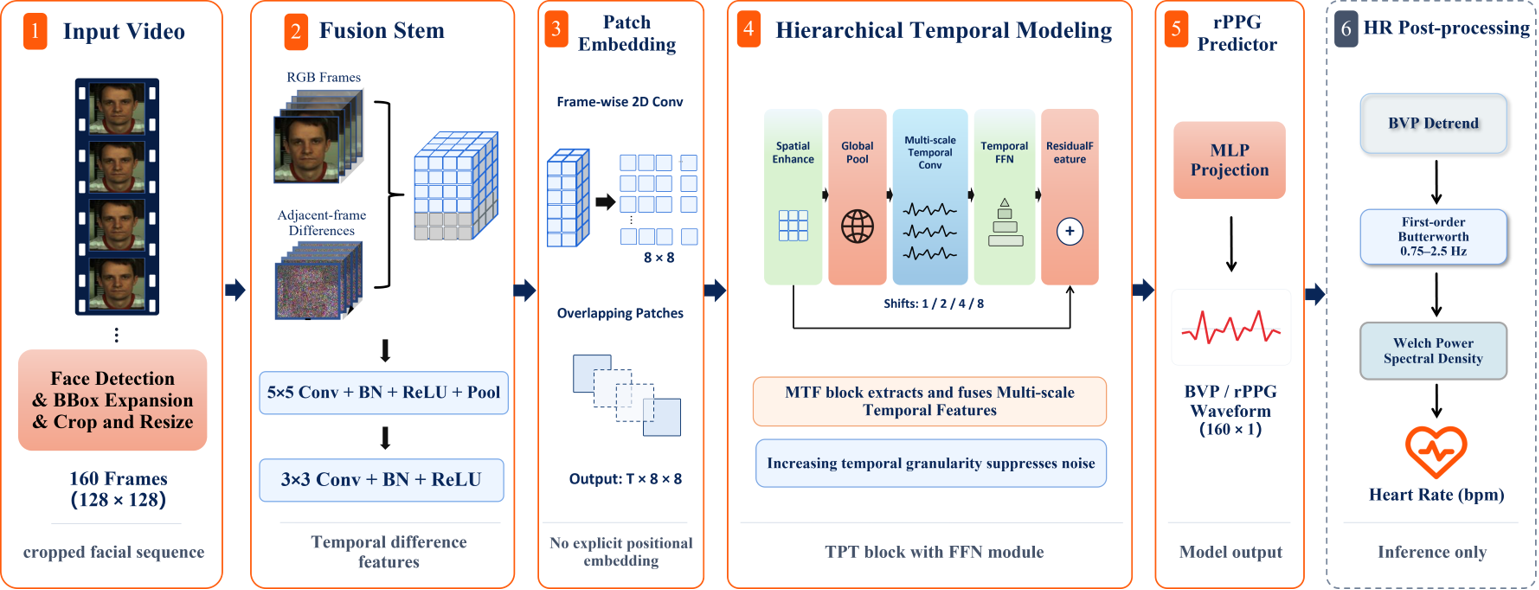}
    \caption{Model architecture of TinyHR.}
    \label{fig:tinyhr_architecture}
\end{figure}

\subsubparagraph{Frame Difference Fusion Stem}
The 12-channel difference representation is first processed by a 2D convolution with a kernel size of 5 and stride 2, followed by batch normalization, ReLU activation, and max pooling with stride 2. A subsequent 3 × 3 convolutional block further refines the feature representation. The resulting feature map contains 16 channels and has one quarter of the original spatial resolution. This design concentrates the early feature extraction process on frame-to-frame variations, which are more directly related to subtle pulsatile color changes in facial skin.

\subsubparagraph{Spatial Patch Embedding}
The output of the Difference Feature Stem is converted into a compact spatial representation on a frame-by-frame basis. A 2D convolution with a kernel size and stride of 4 reduces the spatial resolution from 32 × 32 to 8 × 8 while maintaining 32 feature channels. Batch normalization and ReLU activation are then applied. The temporal dimension is not downsampled during this operation; after processing individual frames, the features are rearranged into a video representation with dimensions corresponding to batch, channel, time, height, and width.

This frame-wise spatial embedding preserves the temporal sequence while reducing the spatial feature grid, providing a compact representation for subsequent spatiotemporal feature extraction.

\subsubparagraph{Multi-Scale Temporal Feature Block}
The core feature extraction module is a Multi-Scale Temporal Feature Block. Unlike an attention-based transformer block, this module models temporal dependencies using convolutional operations and explicit temporal sampling. The model uses a single such feature-extraction stage, without hierarchical temporal downsampling or multiple transformer stages. The block maintains the spatial feature map while deriving a compact temporal representation for physiological signal modeling.

First, a frame-wise spatial enhancement module processes each frame independently. The feature tensor is temporarily rearranged so that the batch and temporal dimensions are combined, and depthwise spatial convolution followed by pointwise convolution is applied to each frame. The enhanced frame features are then restored to the original spatiotemporal layout.

Next, spatial global average pooling aggregates the spatial dimensions of each frame, converting the feature map into a channel-wise temporal sequence. This sequence is processed by a multi-scale temporal convolution module with four parallel branches. The four branches use temporal shifts with different offsets, providing progressively larger temporal contexts. Each branch combines the shifted features with the current feature through channel concatenation and pointwise convolution. The outputs of the four branches are subsequently concatenated and fused by another pointwise convolution, producing a unified multi-scale temporal representation.

The resulting temporal features are further refined by a temporal feed-forward network composed of pointwise convolutional layers, batch normalization, and ReLU activation. The refined temporal response is then broadcast back to the spatial feature map and added to the enhanced spatial representation. A residual connection from the block input is retained, followed by a ReLU activation. Through this process, the block jointly preserves frame-level spatial information and injects multi-scale temporal physiological cues into the feature representation.

\subsubparagraph{Temporal Feature Feed-Forward Network}
The temporal feed-forward network is used to increase the nonlinear representation capacity of the temporal features after multi-scale temporal aggregation. It expands the channel dimension, applies a sequence of pointwise convolution, batch normalization, and ReLU operations, and projects the representation back to the original channel dimension. A residual connection is used within the module so that the refined temporal representation retains information from the input sequence.

\subsubparagraph{Waveform Predictor}
After multi-scale temporal feature extraction, spatial global average pooling produces a compact temporal representation. The waveform predictor then applies two pointwise convolutional layers to map the temporal features to a single-channel signal. The predicted output is an rPPG waveform with one value for each video frame.

The neural network therefore predicts the rPPG waveform directly rather than the heart rate itself. Heart-rate estimation is performed after waveform prediction using the physiological signal-processing procedure.

\subsubsubsubsection{Training Objectives}
\label{training-objectives}
The network is optimized using complementary time-domain and frequency-domain objectives. The time-domain objective encourages the predicted waveform to reproduce the temporal morphology and phase of the reference BVP signal, while the frequency-domain objectives explicitly constrain the predicted signal toward the correct heart-rate distribution.
Let
$\mathbf{S} = \{ S_{t}\}_{t = 1}^{T}$
denote the reference BVP waveform and
$\widehat{\mathbf{S}} = \{{\widehat{S}}_{t}\}_{t = 1}^{T}$
denote the predicted rPPG waveform for one training sample.

\subsubparagraph{Time-Domain Loss}
\label{time-domain-loss}
The time-domain objective is based on the negative Pearson correlation coefficient. First, the temporal means of the reference and predicted waveforms are defined as
$\mu_{S} = \frac{1}{T}\sum_{t = 1}^{T}S_{t}$ and $\mu_{\widehat{S}} = \frac{1}{T}\sum_{t = 1}^{T}{\widehat{S}}_{t}$. The Pearson correlation coefficient is then formulated as

$$\rho\left( \mathbf{S},\widehat{\mathbf{S}} \right) = \frac{\sum_{t = 1}^{T}\left( S_{t} - \mu_{S} \right)\left( {\widehat{S}}_{t} - \mu_{\widehat{S}} \right)}{\sqrt{\sum_{t = 1}^{T}\left( S_{t} - \mu_{S} \right)^{2}}\sqrt{\sum_{t = 1}^{T}\left( {\widehat{S}}_{t} - \mu_{\widehat{S}} \right)^{2}}}.$$

The time-domain loss is defined as $\mathcal{L}_{time} = - \rho\left( \mathbf{S},\widehat{\mathbf{S}} \right)$.
Because Pearson correlation is invariant to affine scaling, this objective focuses on the temporal morphology and phase consistency between the predicted and reference waveforms rather than their absolute amplitudes.

\subsubparagraph{Frequency-Domain Cross-Entropy Loss}
\label{frequency-domain-cross-entropy-loss}
To explicitly constrain the physiological frequency of the predicted signal, the waveform is transformed into the frequency domain. Let \(f_{s}\) denote the video sampling rate and \(w_{t}\) denote a Hann window. The windowed discrete Fourier transform of the waveform is
$\mathcal{F}(f) = \sum_{t = 1}^{T}w_{t}{\widehat{S}}_{t}e^{- j2\pi ft/f_{s}}$.
The corresponding power spectral density (PSD) is
$P_{\widehat{S}}(f) = \left| \mathcal{F}(f) \right|^{2}$.
Let \(\mathcal{F}_{HR}\) denote the set of physiologically valid frequency bins. The normalized PSD is converted into a categorical probability distribution over heart-rate bins:
\[p_{\widehat{S}}(f) = \frac{P_{\widehat{S}}(f)}{\sum_{f' \in \mathcal{F}_{HR}}^{}P_{\widehat{S}}(f')},\quad\quad f \in \mathcal{F}_{HR}.\]

The reference BVP signal is processed in the same manner to obtain\({\ P}_{S}(f)\). The target heart-rate class is defined as the frequency bin corresponding to the dominant peak of the reference BVP spectrum:
\(f^{*} = \underset{f \in \mathcal{F}_{HR}}{arg\, max}\mspace{6mu} P_{S}(f).\)
Let \(y_{f}\) denote the one-hot target distribution,
$y_{f} = \left\{ \begin{matrix}
1, & f = f^{*}, \\
0, & \text{otherwise}.
\end{matrix} \right.\ $
The frequency-domain cross-entropy loss is therefore

\[\mathcal{L}_{CE} = - \sum_{f \in \mathcal{F}_{HR}}^{}y_{f}\log p_{\widehat{S}}(f).\]

In the official implementation, the cross-entropy path uses heart-rate bins corresponding to 45--149 BPM. This objective complements the time-domain loss by explicitly encouraging the predicted waveform to concentrate its spectral energy around the physiological heart rate of the reference signal.

\subsubparagraph{2.3 Heart-Rate Distribution Constraint}
\label{heart-rate-distribution-constraint}
In addition to the categorical frequency-domain supervision, an HR-aware distribution constraint is introduced to measure the similarity between the spectral distributions of the reference and predicted signals.

Specifically, a Gaussian distribution is constructed around the dominant PSD frequency of each waveform. Let \(f_{S}^{*}\) \emph{and} \(f_{\widehat{S}}^{*}\)denote the dominant frequency bins of the reference and predicted signals, respectively. Their corresponding Gaussian distributions can be written as
\(q_{S}(f) = \frac{1}{Z_{S}}\exp\left( - \frac{\left( f - f_{S}^{*} \right)^{2}}{2\sigma^{2}} \right)\)
and
\(q_{\widehat{S}}(f) = \frac{1}{Z_{\widehat{S}}}\exp\left( - \frac{\left( f - f_{\widehat{S}}^{*} \right)^{2}}{2\sigma^{2}} \right)\),
where \(\sigma\) controls the distribution width and \(Z_{S}\) and \(Z_{\widehat{S}}\) are normalization constants.

The divergence between the reference and predicted HR distributions is measured using the Kullback-Leibler (KL) divergence:
\[\mathcal{L}_{KL} = D_{KL}\left( q_{S}\, \parallel \, q_{\widehat{S}} \right) = \sum_{f \in \mathcal{F}_{HR}}^{}q_{S}(f)\log\frac{q_{S}(f)}{q_{\widehat{S}}(f)}.\]
This constraint provides an HR-oriented measure of the agreement between the predicted and reference spectral distributions.

\subsubparagraph{Overall Training Objective}
\label{overall-training-objective}
The complete training objective combines the time-domain waveform loss and the frequency-domain supervision:
\[\mathcal{L =}\lambda_{time}\mathcal{L}_{time} + \lambda_{CE}\mathcal{L}_{CE} + \lambda_{KL}\mathcal{L}_{KL},\]
where \(\lambda_{time}\)\emph{,} \(\lambda_{CE}\), and \(\lambda_{KL}\) are weighting coefficients set as 0.2, 1.0, 1.0, respectively.
The three terms provide complementary supervision. Specifically, \(\mathcal{L}_{time}\) \emph{encourages temporal waveform consistency,} \(\mathcal{L}_{CE}\) explicitly identifies the correct heart-rate frequency bin, and \(\mathcal{L}_{KL}\) encourages agreement between the predicted and reference HR distributions. Together, they guide the network to generate an rPPG waveform that is simultaneously temporally consistent and physiologically meaningful.

\subsubsubsubsection{Heart-Rate Estimation During Inference}
\label{heart-rate-estimation-during-inference}
The output of the neural network is an rPPG waveform rather than a heart-rate value. Therefore, a deterministic physiological signal-processing pipeline is applied after waveform prediction.
Given the predicted waveform \(\widehat{S}\), a second-order Butterworth band-pass filter with cutoff frequencies of 0.75 Hz and 2.5 Hz is first applied to suppress frequencies outside the expected physiological heart-rate range. The power spectral density is then estimated using Welch's method. The dominant frequency is obtained as

\[f^{*} = \underset{f}{\arg{\max} \; P(f)},\]
where \(P(f)\) denotes the estimated PSD of the filtered rPPG signal. Finally, the heart rate is calculated as
\(HR = 60f^{*}\),
where HR is expressed in beats per minute (BPM) and \(f^{*}\) is measured in Hz.

\subsubsubsubsection{Training Configuration}\label{training-configuration-and-experimental-results}

The model was trained using a combination of multiple rPPG datasets to improve the diversity of subjects, recording conditions, and physiological signals. The training set consisted of four datasets: \textbf{VIPL-HR V1}\cite{niu2018viplhr},\textbf{VIPL-HR V2}, \textbf{V4V}, and \textbf{MCD-rPPG}.
Specifically, VIPL-HR V1 contains 85 subjects and 1,883 videos, while VIPL-HR V2 contains 500 subjects and 2,498 videos. The V4V dataset contains 103 subjects and 726 videos. For MCD-rPPG, only front-facing videos were retained for training, resulting in 600 subjects and 1,200 videos.
The resulting training set therefore covers a relatively broad range of subjects and recording conditions, providing diverse training samples for learning robust rPPG representations.

The model was trained with a batch size of 4. The initial learning rate was set to 0.005, and the \textbf{OneCycleLR} learning-rate scheduling strategy was adopted to dynamically adjust the learning rate throughout training. 

\subsubsection{Performance}
We evaluate the heart-rate estimation methods used in SeetaPsych v1.0, including the unsupervised \textbf{AdaChrom} method and the supervised \textbf{Tiny-HR} model, on the VIPL-HR dataset~\cite{niu2018viplhr}. For AdaChrom, several configurations with different ROI settings are considered according to the ROI definitions introduced in the ROI Mask Generation section. In addition, we further report the performance of AdaChrom and Tiny-HR on the official fifth-fold test set of VIPL-HR, which covers a broader range of recording conditions.

\subparagraph{Evaluation Dataset}
We report the results of different ROI configurations on the VIPL-HR dataset~\cite{niu2018viplhr}, a large-scale benchmark for remote heart-rate estimation from less-constrained facial videos. VIPL-HR contains recordings from 107 subjects and provides synchronized physiological annotations, including heart-rate values and pulse waveform signals. 
\subparagraph{Evaluation Protocol}
We report results using two reference labels, denoted as \textbf{Label-gt} and \textbf{Label-wave}.

For \textbf{Label-gt}, the reference heart rate is directly obtained from the official heart-rate annotations provided by VIPL-HR~\cite{niu2018viplhr}. The estimated heart rate produced by each method is then compared with this reference value. This protocol measures the consistency between the estimated results and the dataset-provided HR annotations.

For \textbf{Label-wave}, the reference heart rate is recomputed from the synchronized BVP waveform provided by VIPL-HR~\cite{niu2018viplhr}. Specifically, the physiological waveform is processed using a consistent heart-rate extraction protocol to obtain the corresponding reference HR, which is then compared with the estimated HR. This protocol reduces dependence on pre-computed HR annotations and provides an additional assessment based directly on the underlying physiological waveform.

\subparagraph{Evaluation Metrics}
We report heart-rate estimation performance using three commonly adopted metrics: mean absolute error (MAE), root mean square error (RMSE), and Pearson correlation coefficient. MAE reflects the average estimation error in BPM, RMSE penalizes larger errors more strongly, and Pearson correlation measures the linear consistency between the estimated and reference heart-rate sequences.

\begin{equation}
\mathrm{MAE} = \frac{1}{N}\sum_{i=1}^{N} \left| \hat{y}_i - y_i \right|,
\end{equation}

\begin{equation}
\mathrm{RMSE} = \sqrt{\frac{1}{N}\sum_{i=1}^{N} \left( \hat{y}_i - y_i \right)^2},
\end{equation}

\begin{equation}
\mathrm{Pearson} =
\frac{
\sum_{i=1}^{N} \left(\hat{y}_i - \bar{\hat{y}}\right)\left(y_i - \bar{y}\right)
}{
\sqrt{\sum_{i=1}^{N} \left(\hat{y}_i - \bar{\hat{y}}\right)^2}
\sqrt{\sum_{i=1}^{N} \left(y_i - \bar{y}\right)^2}
},
\end{equation}
where $\hat{y}_i$ and $y_i$ denote the estimated and reference heart rates of the $i$-th valid evaluation window, respectively, $N$ is the number of valid windows, and $\bar{\hat{y}}$ and $\bar{y}$ are their mean values.

\subparagraph{Experiment I: Effect of ROI Configurations}
We first investigate the effect of different ROI configurations on AdaChrom. For this experiment, we use the \textit{stable} and \textit{long-distance} scenarios of VIPL-HR~\cite{niu2018viplhr}. These two scenarios contain relatively limited head motion and pose variation, making them suitable for isolating the influence of ROI selection on the extracted physiological signal. At the same time, the long-distance setting introduces changes in face scale and signal strength, allowing us to examine the robustness of different ROI configurations under weaker facial observations.

All experiments were conducted on an Intel(R) Core(TM) i7-10750H CPU @ 2.60 GHz, and the results are reported in Table~\ref{tab:AdaChrom}.

\begin{table}[htbp]
    \centering
    \caption{Results on stable and long-distance scenarios of the VIPL-HR dataset~\cite{niu2018viplhr}.}
    \label{tab:AdaChrom}

    \resizebox{\linewidth}{!}{%
    \begin{tabular}{lccccccc}
        \toprule
        \multirow{2}{*}{Method}
        & \multicolumn{3}{c}{Label-gt}
        & \multicolumn{3}{c}{Label-wave}
        & \multirow{2}{*}{\shortstack{Throughput\\(FPS)}} \\
        
        \cmidrule(lr){2-4}
        \cmidrule(lr){5-7}
        
        & MAE$\downarrow$
        & RMSE$\downarrow$
        & Pearson$\uparrow$
        & MAE$\downarrow$
        & RMSE$\downarrow$
        & Pearson$\uparrow$
        & \\
        
        \midrule
        
        AdaChrom-v1
        & 5.77 & 8.99 & 0.64
        & 3.99 & 6.27 & 0.75
        & 131 \\
        
        AdaChrom-v2
        & 5.70 & 8.75 & 0.68
        & 3.79 & 5.65 & 0.79
        & 99 \\
        
        \textbf{AdaChrom-v3}
        & 5.64 & 8.60 & 0.68
        & 3.66 & 5.44 & 0.81
        & 95 \\
        
        AdaChrom-v4
        & 5.63 & 8.59 & 0.68
        & 3.69 & 5.45 & 0.81
        & 94 \\
        
        \bottomrule
    \end{tabular}%
    }
\end{table}

As shown in Table~\ref{tab:AdaChrom}, increasing the ROI configuration from AdaChrom-v1 to AdaChrom-v3 consistently improves heart-rate estimation accuracy, although the processing throughput decreases accordingly. AdaChrom-v3 achieves an MAE of 3.66 BPM and an RMSE of 5.44 BPM under Label-wave, together with a Pearson correlation of 0.81. AdaChrom-v4 provides nearly identical performance, with only marginal differences under Label-gt and Label-wave, while introducing slightly higher computational cost. We therefore select \textbf{AdaChrom-v3} as the default configuration, as it provides a favorable balance between estimation accuracy and computational efficiency.

\subparagraph{Experiment II: Evaluation on the Fifth Fold of VIPL-HR}
Following the default settings of rPPG-Toolbox~\cite{liu2023rppg}, an open-source benchmark toolbox for remote physiological measurement, we further report the performance of AdaChrom and Tiny-HR on the official fifth-fold test set of the VIPL-HR dataset~\cite{niu2018viplhr}. This test set contains 485 videos from 22 subjects and covers a broader range of recording conditions than the relatively controlled scenarios used in Experiment~I. Therefore, this experiment is used to assess the robustness of the two methods under more diverse test settings.

As an unsupervised method, AdaChrom is directly applied to the test videos without training. For Tiny-HR, the fifth-fold test set is used only for evaluation and is excluded from the training data. The results are reported in Table~\ref{tab:vipl_hr_fifth_fold}. Tiny-HR achieves an MAE of 3.88 BPM on the fifth-fold test set, demonstrating substantially improved robustness under the diverse conditions included in the full VIPL-HR evaluation split.

\begin{table}[htbp]
    \centering
    \caption{Results on the fifth fold of the VIPL-HR dataset~\cite{niu2018viplhr}.}
    \label{tab:vipl_hr_fifth_fold}

    \resizebox{0.8\linewidth}{!}{%
    \begin{tabular}{lcccccc}
        \toprule
        \multirow{2}{*}{Method}
        & \multicolumn{3}{c}{Label-gt}
        & \multicolumn{3}{c}{Label-wave} \\
        
        \cmidrule(lr){2-4}
        \cmidrule(lr){5-7}
        
        & MAE$\downarrow$
        & RMSE$\downarrow$
        & Pearson$\uparrow$
        & MAE$\downarrow$
        & RMSE$\downarrow$
        & Pearson$\uparrow$ \\
        
        \midrule
        
        AdaChrom-v3
        & 8.60 & 13.00 & 0.42
        & 6.63 & 9.84 & 0.47 \\

        Tiny-HR
        & 5.22 & 8.68 & 0.68
        & 3.88 & 6.89 & 0.77 \\

        \bottomrule
    \end{tabular}%
    }
\end{table}

\subparagraph{Discussion}
The two experiments highlight the complementary characteristics of AdaChrom and Tiny-HR. AdaChrom is a training-free method that performs well when the face is relatively stable and the selected ROI contains reliable physiological signals. Its performance decreases on the more diverse fifth-fold test set, where variations in motion, pose, illumination appearance, and acquisition conditions make purely signal-based estimation more challenging. In contrast, Tiny-HR benefits from supervised learning on labeled data and can better model these variations, leading to stronger accuracy under more challenging and unconstrained conditions. Therefore, in SeetaPsych v1.0, the two methods provide different trade-offs among computational efficiency, deployment flexibility, and robustness to complex recording conditions.

\subsection{Screen point-of-gaze estimation}

Screen point-of-gaze estimation provides a direct measure of where a user is visually attending on a display, making it useful for studying visual attention and human–computer interaction. SeetaPsych v1.0 provides two complementary solutions for this task: a direct 2D solution that estimates gaze position from visual appearance and facial geometry, and a 3D-gaze-based solution that predicts the 3D gaze ray, represented by its origin and direction, and obtains the point-of-gaze by intersecting it with the screen plane. This section describes the formulation and implementation of these two complementary solutions.

\subsubsection{Functional Description}

The screen point-of-gaze estimation module predicts the 2D location on a screen at which a user is looking from an image captured by a screen-mounted camera.

\subsubsection{Input and Output}
\textbf{Input:}

\begin{adjustwidth}{4em}{0em}
\textbf{Camera image:}
An RGB image captured by a camera mounted on the screen.
\end{adjustwidth}

\textbf{Outputs:}
{
\setlength{\leftmargini}{4em}
\begin{enumerate}
    \renewcommand{\labelenumi}{\arabic{enumi}.}
    \setlength{\itemsep}{0.5em}
    \setlength{\topsep}{0.4em}

    \item \textbf{Screen point of gaze (physical coordinates):}
    A 2D coordinate
    \(
    (x, y)
    \)
    representing the estimated gaze position in physical screen coordinates,
    expressed in millimetres (mm) for the left or right eye. For some algorithms, the gaze of two eye are considered the same.
    The coordinate system is illustrated in Figure~\ref{fig:pog_coordinate_system}.

    \begin{figure}[htbp]
    \centering
    \includegraphics[width=0.2\linewidth]{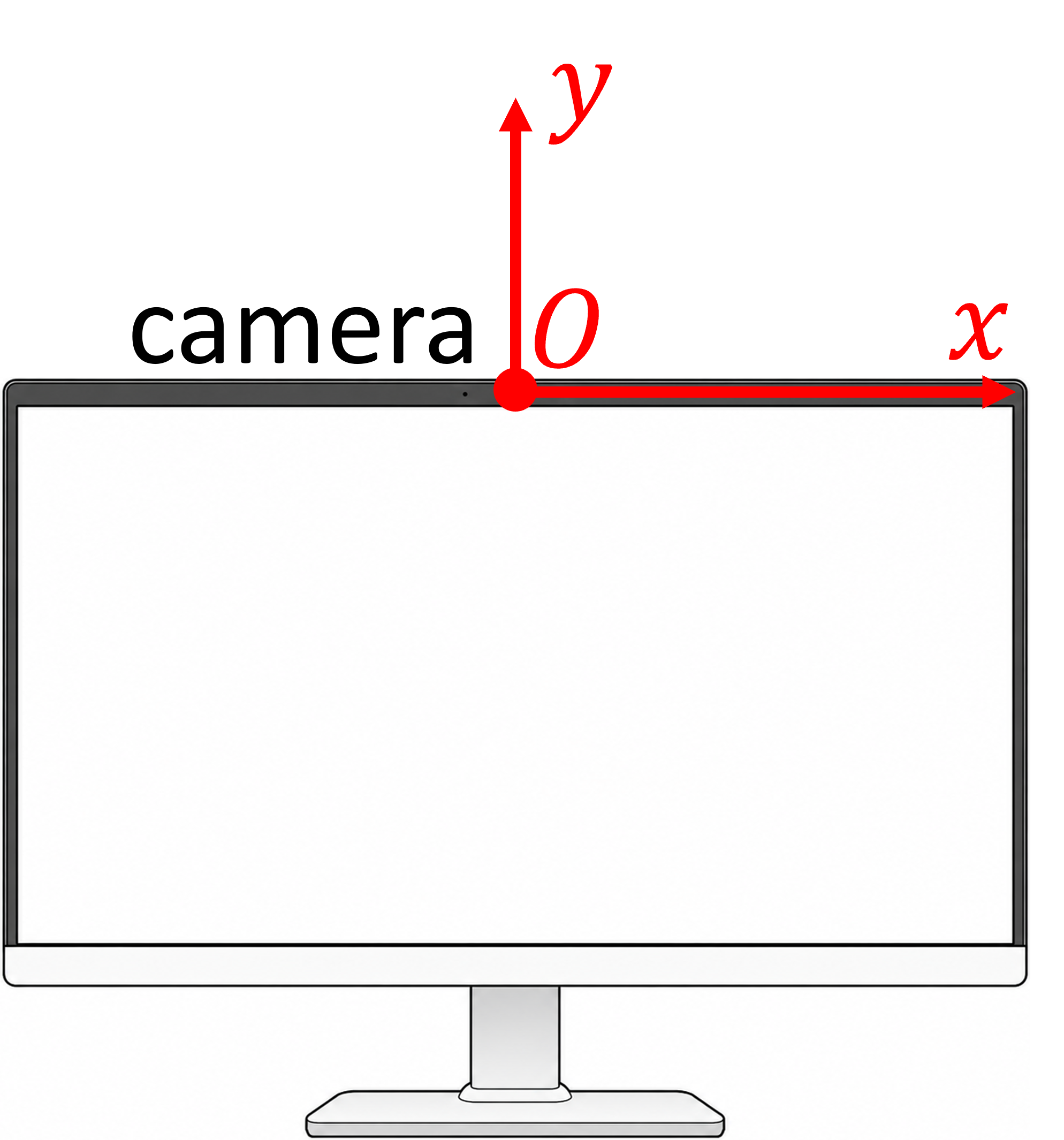}
    \caption{The physical 2D coordinate system for the Point-of-Gaze on the screen (Unit: mm).}
    \label{fig:pog_coordinate_system}
\end{figure}

    \item \textbf{Screen point of gaze (pixel coordinates):}
    A 2D coordinate
    \(
    (u, v)
    \)
    representing the corresponding gaze position in screen pixel coordinates,
    expressed in pixels (px) for the left or right eye. For some algorithms, the gaze of two eye are considered the same.
    The coordinate system is illustrated in Figure~\ref{fig:pog_coordinate_system_px}.

    \begin{figure}[htbp]
    \centering
    \includegraphics[width=0.25\linewidth]{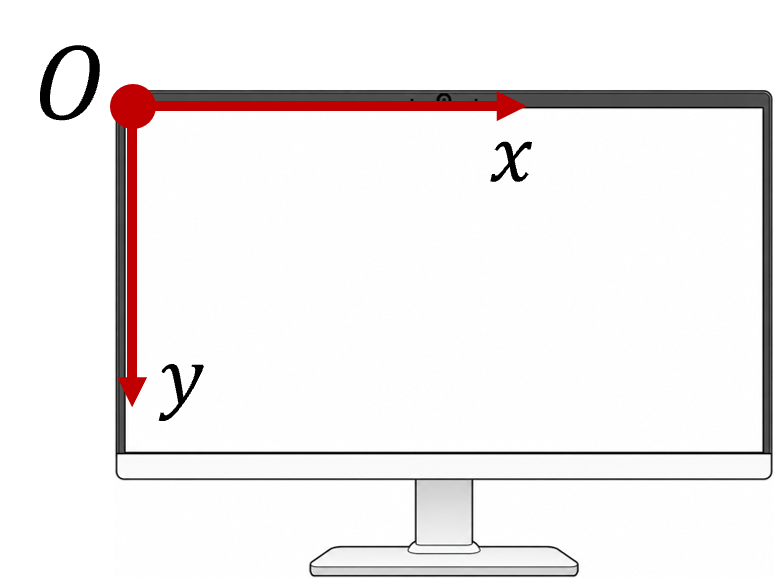}
    \caption{The pixel 2D coordinate system for the Point-of-Gaze on the screen (Unit: pixel).}
    \label{fig:pog_coordinate_system_px}
\end{figure}

\end{enumerate}
}

\subsubsection{Core Algorithms and Models}
The screen point-of-gaze estimation module provides two pipelines. The direct 2D approach regresses screen gaze position from facial and eye appearances together with facial geometry, with trained models based on AFFNet\cite{bao2021affnet}. The 3D gaze-based approach estimates the gaze origin and direction using a trained TDGazeNet-based model. The screen point of gaze is then obtained by intersecting the predicted gaze ray with the screen plane. The following subsections describe both pipelines in detail.

\begin{figure}[htbp]
    \centering
    \includegraphics[width=1\linewidth]{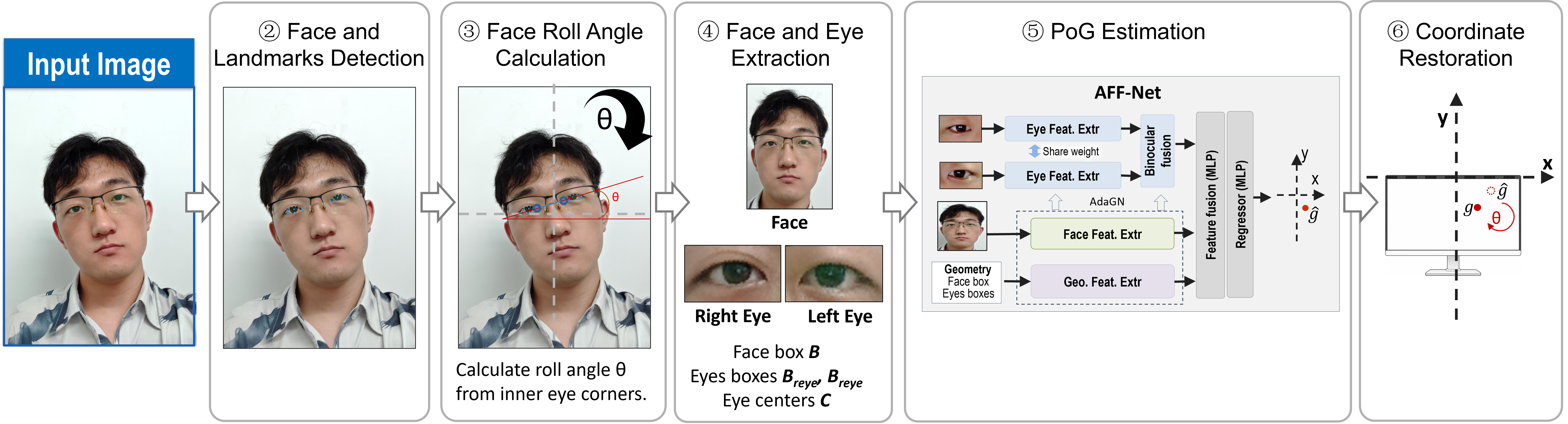}
    \caption{Pipeline for Direct 2D Point-of-Gaze Estimation.}
    \label{fig:direct_pog_pipeline}
\end{figure}

\subsubsubsection{Direct 2D Point-of-Gaze Estimation}
This solution directly predicts the 2D point-of-gaze on the screen. Figure~\ref{fig:direct_pog_pipeline} shows the processing pipeline that consists of the following steps.

\subsubsubsubsection{Face and landmark detection} 
Given an input image, this solution first applies face detection and facial landmark detection tools to obtain the facial landmarks. The detected landmarks are then used to locate the image coordinates of the inner corners of the two eyes.

\subsubsubsubsection{In-plane image rotation} This step is introduced to reduce the variation caused by in-plane head rotation. The front-camera image used in this solution is treated as the mirrored version of the image formed by the camera under the pinhole-camera model. Under this imaging geometry, an in-plane roll of the face by a given angle corresponds to an opposite rotation of the screen PoG around the coordinate origin. This geometric correspondence allows us to normalize the facial roll by rotating the input image while preserving the ability to recover the original PoG through the inverse transformation. Therefore, the gaze estimation model mainly focuses on learning gaze variations under the normalized facial pose, reducing the impact of in-plane head rotation.

Specifically, the in-plane face rotation angle \(\theta\) is computed as
the angle between the line connecting the two inner eye corners and the
horizontal image axis. The input image and facial landmarks are then rotated around the image center by \(\theta\) to obtain an approximately
roll-normalized face image.

\subsubsubsubsection{Face and eye region cropping}
 Based on the transformed landmarks, three regions are cropped: the face image \(I_{face}\), the right-eye image \(I_{reye}\), and the left-eye image \(I_{leye}\). The transformed landmarks are also used to derive the face bounding box and the two eye-center coordinates, which provide additional facial geometry information.

\subsubsubsubsection{PoG estimation} The normalized face and eye regions, together with the facial geometry information, are fed into an end-to-end gaze estimation model to directly regress the screen PoG, denoted as $\hat{g}$. SeetaPsych provides model weights with open-sourced framework AFFNet~\cite{bao2021affnet}. The model is trained on the union of the GazeCapture dataset and additional self-collected gaze data.

\subsubsubsubsection{Coordinate restoration} 
The predicted gaze point \(\hat{g}\) is obtained from the roll-normalized
image and is therefore expressed in the corresponding normalized screen
coordinate system. To recover the gaze point corresponding to the original image before roll normalization, \(\hat{g}\) is transformed using the same in-plane rotation applied during preprocessing, with the rotation centered at the coordinate origin corresponding to the camera position as shown in Figure~\ref{fig:pog_coordinate_system}. For example, if the original image is rotated clockwise by an angle \(\theta\) during preprocessing, \(\hat{g}\) is rotated clockwise by the same angle \(\theta\) around the coordinate origin to obtain the final gaze point \(g\).

The resulting screen PoG $g = (x, y)$ is provided in physical screen coordinates (Fig.~\ref{fig:pog_coordinate_system}),
expressed in millimetres.
The toolbox also converts $g$ from physical coordinates to screen pixel coordinates (Fig.~\ref{fig:pog_coordinate_system_px}).
The final screen PoG is therefore provided in both physical screen
coordinates \((x, y)\) in millimetres and screen pixel coordinates
\((u, v)\) in pixels.

\subsubsubsection{3D Gaze-Based Point-of-Gaze Estimation}

This solution first estimates the 3D gaze origin and gaze direction and then computes the screen PoG through geometric gaze-ray intersection. The processing pipeline consists of the following steps.

\begin{figure}[htbp]
    \centering
    \includegraphics[width=1\linewidth]{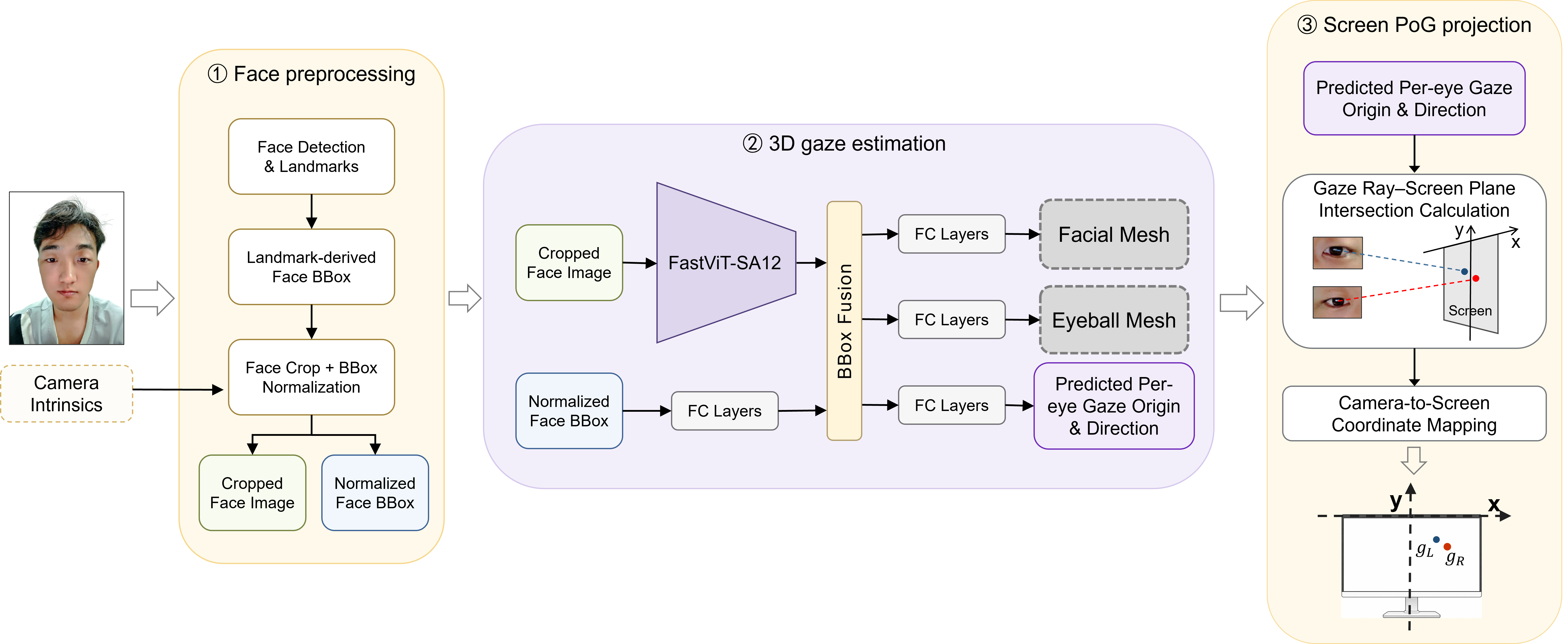}
    \caption{Pipeline for 3D Gaze-Based Point-of-Gaze Estimation.}
    \label{fig:3d_pog_pipeline}
\end{figure}

\subsubsubsubsection{Face preprocessing}
Given an input image, this solution first corrects lens distortion using the camera intrinsics and distortion coefficients, which can be obtained in advance using a standard camera calibration procedure such as Zhang's method~\cite{zhang2000flexible}. Facial landmarks are then detected, from which a face bounding box is derived. Let \(w\) and \(h\) denote its width and height. A square crop centred at the bounding-box centre is defined with side length \(s=\alpha\sqrt{wh}\), where \(\alpha\) is a configurable scale factor set to \(1.1\) by default. The resulting face region is cropped and resized to \(224 \times 224\) for model input. In parallel, the corresponding square bounding box is mapped from the input-camera image plane to a canonical image plane using the camera intrinsic matrices, following the normalization procedure in \cite{zhang2018revisiting}. The normalized bounding box is then used to encode the spatial position and scale of the face in the original camera view.

\subsubsubsubsection{3D gaze estimation} The \(224 \times 224\) face crop and the normalized face bounding box are fed into TdGazeNet. The network jointly predicts a facial mesh, per-eye eyeball meshes, and the corresponding per-eye gaze origins and directions in the camera coordinate system.

TdGazeNet is developed based on 3DGazeNet~\cite{ververas20243dgazenet} with four major modifications. First, it uses a high-resolution face crop without separate eye crops to preserve fine-grained eye appearance while retaining contextual facial information. Second, the normalized face bounding box is introduced to encode the spatial position and scale of the head, compensating for spatial information lost during face cropping and improving 3D structure estimation under perspective projection. Third, a multi-task regression module is adopted to coordinate the optimization of multiple prediction targets. Finally, FastViT-SA12~\cite{vasu2023fastvit} is used as the backbone, where shallow large-kernel convolutions capture local eye features and deeper self-attention layers model global facial and head-pose information.

TdGazeNet is trained on a synthetic dataset constructed using parametric 3D face modeling and physics-based rendering techniques.

\subsubsubsubsection{Screen PoG projection}

The point of gaze on the screen can be computed as the intersection of the gaze ray with the screen plane. TDGazeNet predicts the gaze origin and gaze direction for both eyes. Without loss of generality, we describe the computation for one eye, while the other eye is handled analogously. Let $\mathbf{o}^c$ and $\mathbf{d}^c$ denote the gaze origin and gaze direction predicted by TDGazeNet in the camera coordinate system, respectively. The corresponding gaze ray can be parameterized as
\[
\mathbf{p}^c(t)=\mathbf{o}^c+t\mathbf{d}^c.
\]

\begin{figure}[htbp]
    \centering
    \includegraphics[width=0.3\linewidth]{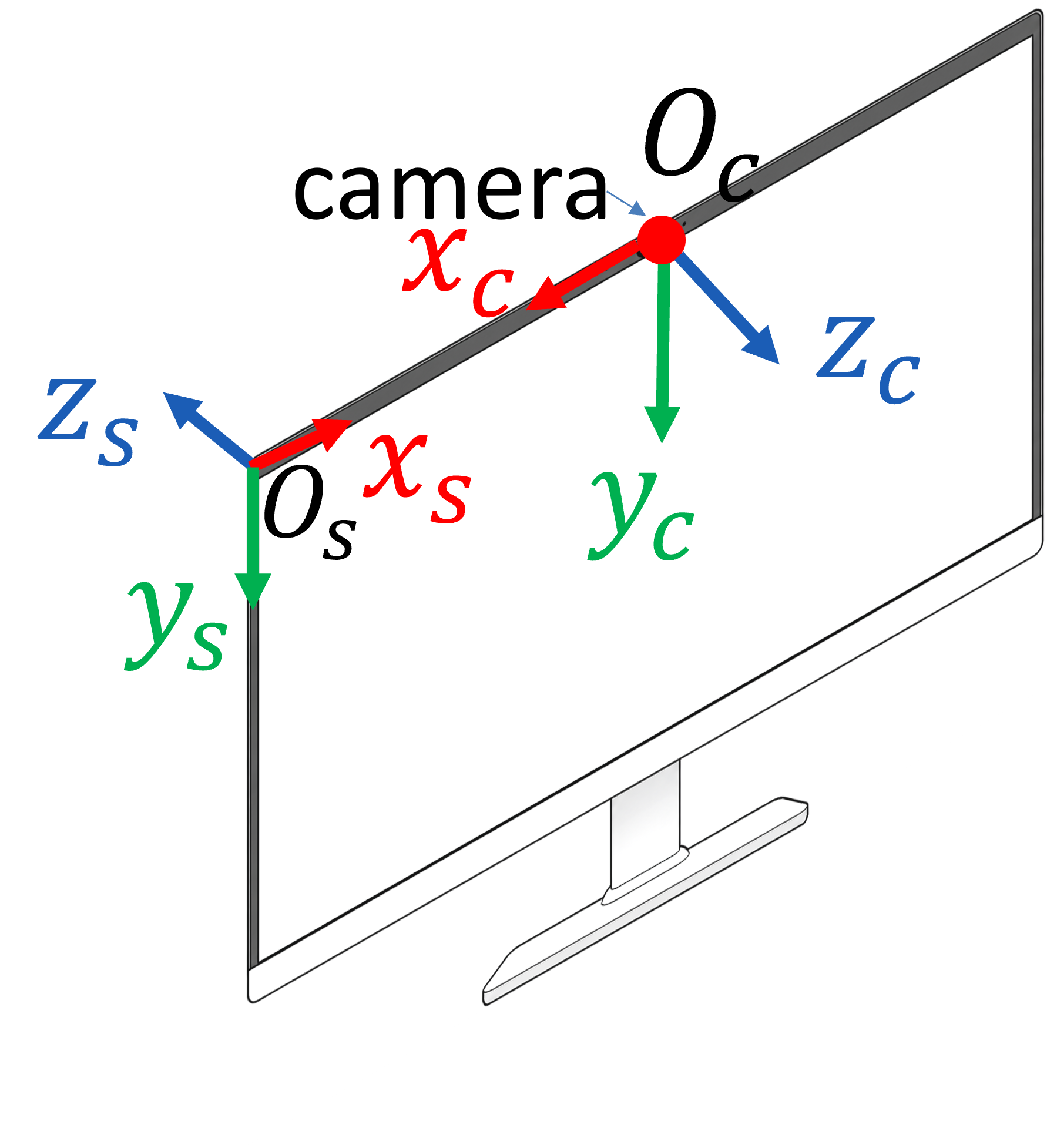}
    \caption{Camera coordinate system and screen coordinate system defined in 3D gaze-based point-of-gaze estimation.}
    \label{fig:3d_coordinate}
\end{figure}

Figure~\ref{fig:3d_coordinate} illustrates the camera coordinate system, defined by the axes $X_c$, $Y_c$, and $Z_c$, and the screen coordinate system, defined by $X_s$, $Y_s$, and $Z_s$. For an arbitrarily positioned camera, the rigid transformation from the camera coordinate system to the screen coordinate system can be obtained through screen--camera calibration, for example, using a mirror-reflection-based method~\cite{francken2007screen}. Let $R$ and $T$ denote the corresponding rotation matrix and translation vector, such that
\[
\mathbf{p}^s = R\mathbf{p}^c + T.
\]

The screen surface corresponds to the $X_s\!-\!O_s\!-\!Y_s$ plane in the screen coordinate system and therefore satisfies $z_s=0$. Accordingly, the screen plane can be expressed in the camera coordinate system as
\[
R_{3,:}\left(\mathbf{p}^c+R^\top T\right)=0,
\]
where $R_{3,:}$ denotes the third row of $R$, and $\mathbf{p}^c$ denotes an arbitrary point on the screen expressed in the camera coordinate system.

Since the point of gaze on the screen lies on both the gaze ray and the screen plane, we have
\[
R_{3,:}
\left(
\mathbf{o}^c+t\mathbf{d}^c+R^\top T
\right)=0.
\]

Solving the above equation yields the intersection parameter $t^*$. The resulting point of gaze in the camera coordinate system is
\[
\mathbf{p}_g^c
=
\mathbf{o}^c+t^*\mathbf{d}^c.
\]

Its coordinates in the screen coordinate system are then computed as
\[
\mathbf{p}_g^s
=
R\left(\mathbf{o}^c+t^*\mathbf{d}^c\right)+T.
\]

For convenient configuration, SeetaPsych also supports the common setup in which the camera is mounted above the screen and the screen plane is approximately parallel to the $X_c\!-\!O_c\!-\!Y_c$ plane of the camera coordinate system. Under the coordinate convention illustrated in Fig.~\ref{fig:3d_coordinate}, the rotation matrix can be set to
\[
R=
\begin{bmatrix}
-1 & 0 & 0\\
0 & 1 & 0\\
0 & 0 & -1
\end{bmatrix}.
\]

The translation vector $T$ is then determined by the relative position between the camera origin $O_c$ and the screen-coordinate origin $O_s$ according to the transformation defined above.

The projection procedure is applied independently to the left and right eyes, yielding two per-eye screen PoG estimates in physical screen coordinates expressed in millimetres. 
Similar to the direct 2D solution, these physical coordinates are also converted to the corresponding screen pixel coordinates.

\subsubsection{Performance}

We evaluate the gaze estimation performance of the two models on an independent self-collected dataset. The evaluation data, metrics, and results are described below.

\subparagraph{Evaluation data} The evaluation set consists of gaze data collected from 58 adult participants who were not included in the training set. Participants were instructed to look at target points displayed on a laptop screen, with the camera mounted directly above the screen. The viewing distance ranged from 30 to 70 cm. The target points covered a horizontal range of $-15$ to $+15$ cm and a vertical range of 0 to 20 cm on the screen.

\subparagraph{Metrics} We report the average and minimum per-subject mean error. 
Specifically, for participant $k$, the mean gaze error $\bar{e}_k$ is defined as the average per-sample gaze error over all test samples of that participant.
It is computed as $$\bar{e}_p = \frac{1}{N_k}\sum_{i=1}^{N_k}||\mathbf{p}_k^{i} - \mathbf{p}_k^{i}||_2,$$
where $\mathbf{p}_k^{i}$ and $\mathbf{p}_k^{i}$ denote the predicted and ground-truth gaze positions of the $i$-th sample, respectively, and $N_k$ is the number of samples for participant $k$. 
Then, the average and minimum of the per-subject mean errors is computed as 
$$\bar{e}_{\text{avg}} = \frac{1}{K}\sum_{k=1}^{K} \bar{e}_k \qquad
\bar{e}_{\text{min}} = \min_k \bar{e}_k,$$ 
where $K$ is the number of participant.

\subparagraph{Results and discussion} 
Table~\ref{tab:gaze_error} summarizes the gaze estimation errors of the two models. Both models show a substantial gap between the average and minimum per-subject errors, indicating considerable inter-subject variation in gaze estimation accuracy.

This variation can be attributed to two main factors. First, the evaluation data were collected under conditions different from those of the training data, introducing a domain shift between training and testing. Second, the models were evaluated without subject-specific calibration. Individual differences in head pose, visual and optical axes (e.g., the kappa angle), and eyeglass prescriptions can therefore introduce systematic subject-dependent errors.

Under these cross-domain and uncalibrated conditions, the average errors remain relatively high, whereas the best-performing participants achieve mean errors of approximately 2-3 cm. This result suggests that model accuracy is strongly affected by individual differences.

\begin{table}[htbp]
    \centering
    \caption{Gaze estimation errors for different models.}
    \label{tab:gaze_error}
    \begin{tabular}{llcc}
        \toprule
        Aggregation & Statistic &  AFFNet & TDGazeNet \\
        \midrule
        \multirow{2}{*}{Per-subject mean error (cm)}
            & Average &  5.82 & 7.75 \\
            & Minimal &  2.35 & 2.29 \\
        \bottomrule
    \end{tabular}
\end{table}

\subsection{Scene gaze following}

Unlike screen point-of-gaze estimation, scene gaze following estimates where a person is looking in a third-person-view scene. SeetaPsych v1.0 adopts CoSI-Gaze to predict gaze targets and, in two-person scenes, further analyze social gaze interactions through contextual and spatial integration. This section describes the formulation and implementation of these functions.

\subsubsection{Functional Description}

The scene gaze following module estimates where a target person is looking in a third-person-view scene image. Given the scene image and the person’s head bounding box, it predicts a gaze-target heatmap and the corresponding gaze point in image pixel coordinates. When two people are present in the scene, the module additionally analyzes their gaze interaction and predicts a social gaze label from five atomic categories: Share, Mutual, Single, Miss, or Void.

\subsubsection{Input and Output}

\textbf{Inputs:}
{
\setlength{\leftmargini}{5em}
\begin{enumerate}
    \renewcommand{\labelenumi}{\arabic{enumi}.}
    \setlength{\itemsep}{0.5em}
    \setlength{\topsep}{0.4em}

    \item \textbf{Scene image:}
    An RGB image provided as a file path, a PIL image, or a NumPy array.

    \item \textbf{Head bounding box:}
    The bounding box of the target person in the original-image pixel coordinates,
    represented as
    \(
        [x_{\min},\, y_{\min},\, x_{\max},\, y_{\max}]
    \).

\end{enumerate}
}

\begin{figure}[htbp]
    \centering
    \includegraphics[width=0.9\linewidth]{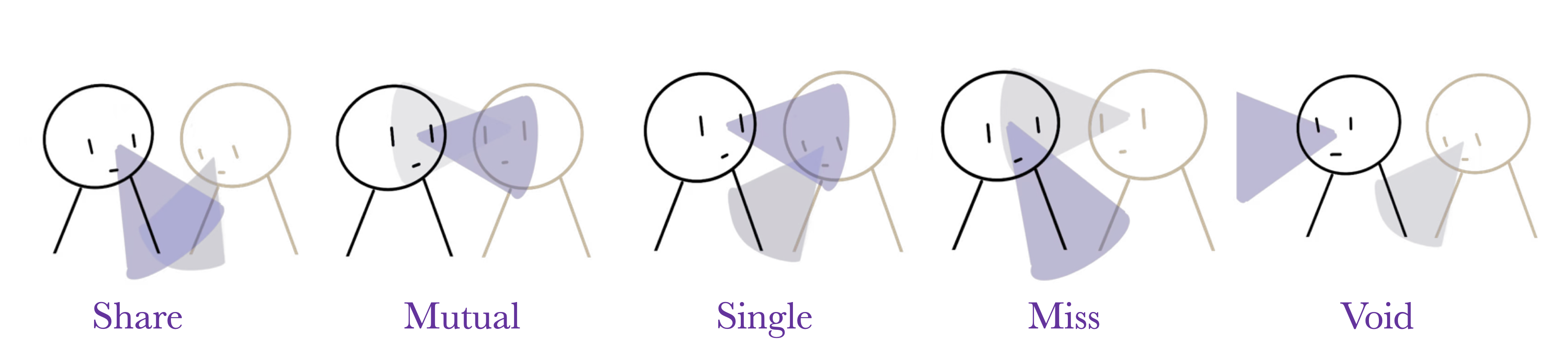}
    \caption{Examples of the five atomic social gaze categories for the principal person, highlighted in a darker color.}
    \label{fig:social_gaze_pattern}
\end{figure}

\textbf{Outputs:}
{
\setlength{\leftmargini}{5em}
\begin{enumerate}
    \renewcommand{\labelenumi}{\arabic{enumi}.}
    \setlength{\itemsep}{0.5em}
    \setlength{\topsep}{0.4em}

    \item \textbf{Gaze heatmap:}
    A single-channel probability map \(G\) over the image plane.
    Each value \(G(x,y)\) indicates the likelihood that location \((x,y)\)
    is the person's gaze target.

    \item \textbf{Gaze point:}
    A 2D coordinate \((\hat{x}, \hat{y})\) corresponding to the location
    with the highest probability in the predicted gaze heatmap \(G\).
    It therefore represents the estimated gaze target location.

    \item \textbf{Social gaze label:}
    If head bounding boxes of two people are provided and the principal person is assigned (the first person by default), the module additionally outputs probabilities over the five atomic social gaze categories defined in~\cite{chang2023gaze}: `Share', `Mutual', `Single', `Miss', and `Void', for the principal person in a dyadic interaction scene. The five categories are illustrated in Figure~\ref{fig:social_gaze_pattern}, where the principal person is highlighted in a darker color.

\end{enumerate}
}

\subsubsection{Core Algorithms and Models}
The implementation of scene gaze following module adopts CoSI-Gaze~\cite{chang2026cosigaze}, a unified framework for gaze following and social gaze prediction. Its gaze-following design is further inspired by Gaze-LLE~\cite{ryan2025gazelle}. CoSI-Gaze combines scene-level visual features with explicit head-position prompts and uses a Transformer-based gaze decoder to produce person-specific gaze representations. For a single person, the module decodes their gaze representations into gaze heatmaps for gaze following. When two people are provided, the module additionally integrates their gaze representations to infer the social gaze labels between them.

\begin{figure}[htbp]
    \centering
    \includegraphics[width=0.95\linewidth]{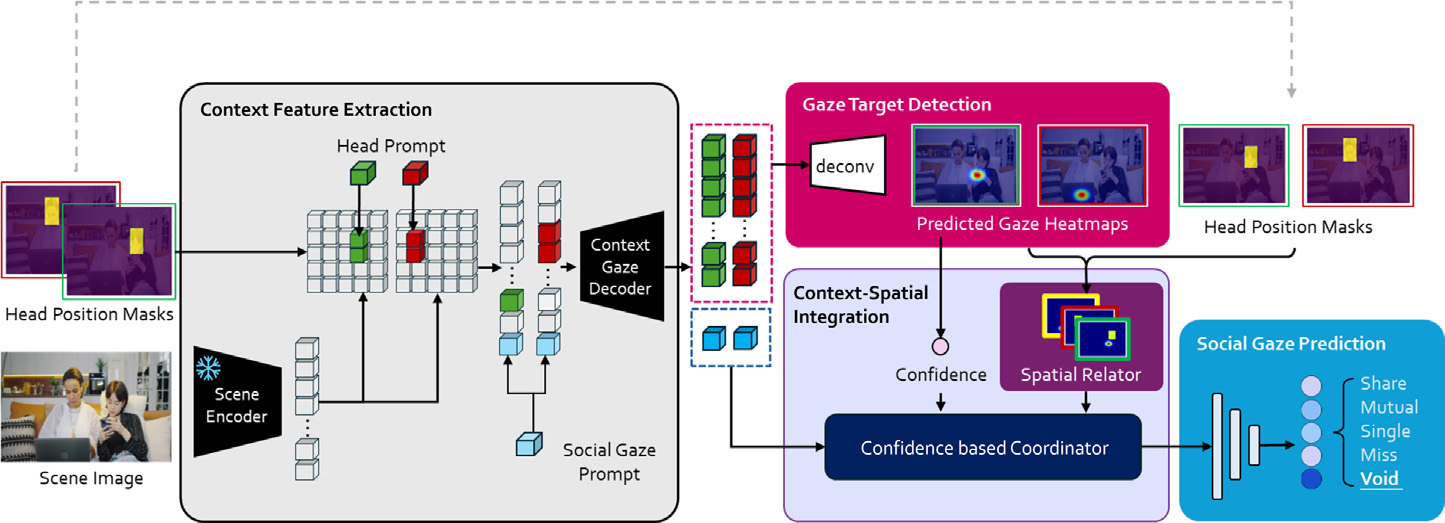}
    \caption{CoSI-Gaze architecture.}
    \label{fig:cosi_gaze_architecture}
\end{figure}

\subparagraph{Pipeline Overview} The gaze-following pipeline consists of two main components: a context feature extractor and a gaze target detection head. The context feature extractor combines scene features with person-specific head information and produces spatial features for gaze target localization. When two people are provided, the social gaze pipeline further uses their predicted gaze heatmaps, head-position masks, and contextual gaze representations to perform context-spatial integration and social gaze classification. The pipeline proceeds as follows:

\subsubparagraph{Scene feature extraction}: The scene encoder processes the scene image and encodes it into a set of feature patches that capture relevant environmental information.
\subsubparagraph{Head position encoding:} Using the head bounding box, the gaze-following module generates a head position mask to indicate the person\textquotesingle s location within the scene.
\subsubparagraph{Head prompt integration:} The module learns a head prompt, which is assigned to the spatial locations indicated by the head position mask. Specifically, the head prompt is added element-wise to the corresponding scene feature patches at the masked locations.
\subsubparagraph{Context gaze decoding:} The context gaze decoder receives the combined features after head prompt integration. It leverages self-attention to model the interactions between person-specific information and scene-level context. It produces refined gaze features.
\subsubparagraph{Gaze target prediction:} The gaze target detection head uses deconvolution layers to transform the decoded features into a gaze heatmap which represents the probability distribution of where the person is looking in the scene.
\subsubparagraph{Social gaze prediction:} When two head bounding boxes are provided, the predicted gaze heatmaps and head-position masks of both people are processed by a Spatial Relator to capture their cross-person spatial relationships. In addition, the module learns a social gaze prompt, which serves as a learnable representation for social gaze reasoning and is used to generate contextual features. The spatial features produced by the Spatial Relator are then combined with these contextual features through a confidence-based coordinator, which adjusts the contribution of spatial information according to the reliability of the gaze predictions. The fused representation is then passed to the social gaze prediction head to classify the dyadic social gaze relationship.

\subparagraph{Model Configuration:} The scene gaze-following module contains 92.57 M parameters in total. It consists of  a DINOv2 scene encoder, a Transformer-based context gaze decoder, a gaze target detection head with transposed convolutional layers and a social-gaze prediction
branch which includes the context-spatial relator and prediction head.

\begin{table}[htbp]
    \centering
    \caption{Model configuration for the main components, including the scene encoder, context gaze decoder, gaze target detection head.}
    \label{tab:gaze_following_configuration}

    \resizebox{\linewidth}{!}{
    \begin{tabular}{lcccc}
        \toprule
        \textbf{Parameter}
        & \textbf{Scene Encoder}
        & \textbf{Context Gaze Decoder}
        & \textbf{Gaze Target Detection Head} 
        & \textbf{Social-Gaze Prediction Branch}\\
        \midrule

        \textbf{Architecture}
        & DINOv2 ViT-B/14
        & Vision Transformer
        & ConvTranspose2d layers 
        & CNN and MLPs \\

        \textbf{Input tensor}
        & $B \times 3 \times 448 \times 448$
        & $B \times 768 \times 32 \times 32$
        & $B \times 768 \times 32 \times 32$ 
        & $B \times 3 \times 64 \times 64$ \\

        \textbf{Output tensor}
        & $B \times 768 \times 32 \times 32$
        & $B \times 768 \times 32 \times 32$
        & $B \times 64 \times 64$ 
        & $B \times 5$ \\

        \textbf{Transformer blocks}
        & 12 & 3 & N/A & N/A \\

        \textbf{Attention heads}
        & 12 & 8 & N/A & N/A \\

        \textbf{Dimension per head}
        & 64 & 32 & N/A & N/A \\

        \textbf{MLP ratio}
        & 4 & 4 & N/A & N/A \\

        \textbf{MLP hidden dimension}
        & 3072 & 1024 & N/A & 512 \\

        \textbf{Activation}
        & GELU & GELU & Sigmoid (heatmap) & RELU\\

        \textbf{QKV / FFN bias}
        & Enabled & Disabled (QKV) & N/A & N/A \\

        \textbf{LayerNorm epsilon}
        & $1 \times 10^{-6}$
        & $1 \times 10^{-6}$
        & N/A & N/A \\

        \textbf{LayerScale initialization}
        & 1.0 & 1.0 & N/A & N/A \\

        \textbf{Kernel size}
        & N/A & N/A & $2 \times 2$ & $3 \times 3$ \\

        \textbf{Stride}
        & N/A & N/A & 2  & 2 \\

        \textbf{Parameter count}
        & 86.58 M & 2.56 M & 0.30 M & 3.13 M\\

        \bottomrule
    \end{tabular}
    }
\end{table}

\subparagraph{Traning Data and Optimization} The gaze-following module is trained using two datasets: GazeFollow~\cite{recasens2015gazefollow} and DyGaze~\cite{chang2026cosigaze}. GazeFollow provides broad scene diversity and contains 122,145 images with 130,339 annotated person instances. DyGaze focuses on natural two-person interactions and contains 95,692 frames with 191,384 person-level annotations from 325 video clips. During training, standard data augmentation is applied, including random cropping, horizontal flipping, and head bounding-box jittering.

Training follows the three-stage strategy of CoSI-Gaze. In the first stage, the context gaze decoder and gaze following head are pretrained on GazeFollow for 15 epochs with a learning rate of $1\times10^{-3}$, followed by 5 epochs on DyGaze with a learning rate of $1\times10^{-4}$. During this stage, the groundtruth gaze locations are converted into two-dimensional Gaussian heatmaps. The module is optimized using mean-squared error (MSE) loss on gaze heatmaps. 

In the second stage, the model is jointly optimized for gaze following and social gaze prediction on DyGaze for additional 5 epochs with a learning rate of $1\times10^{-3}$. Gaze heatmap MSE loss and the social gaze classification loss are jointly applied. During this stage, the context-spatial integration module is bypassed, and social gaze is predicted directly from the learned contextual representation.

In the final stage, the previously trained components are frozen, and the context-spatial integration module together with the social gaze prediction head is trained on DyGaze for 5 epochs with a learning rate of $1\times10^{-4}$. This stage learns to combine contextual and spatial gaze cues for social gaze prediction.

\subsubsection{Performance}

\subparagraph{Evaluation Data} The gaze-following module is evaluated on the DyGaze test set. DyGaze contains videos on natural two-person interactions and provides pixel-level gaze targets. The full dataset contains 95,692 frames and 191,384 person instances; the held-out test partition contains 39,186 person-level samples.
\subparagraph{Evaluation Metrics} Gaze following is evaluated using two metrics: Area Under the Curve (AUC) and L2 distance. AUC evaluates the quality of the predicted gaze heatmap by comparing it with the ground-truth gaze location and computing the area under the ROC curve. L2 distance measures the Euclidean distance between the predicted gaze point and the ground-truth gaze point. Social gaze prediction is evaluated using precision, recall, and F1-score for each social gaze class.

\subparagraph{Compared Methods} The compared methods include VideoAtt, Sharingan and ViTGaze.
\textbf{VideoAtt}~\cite{chong2020videoatt} uses a two-stream CNN architecture that separately encodes the scene and cropped head, followed by deconvolutional layers for gaze prediction.
\textbf{Sharingan}~\cite{tafasca2024sharingan} uses a transformer-based architecture that represents each person with a person-specific gaze token and jointly processes these tokens with image tokens, followed by a conditional multi-scale DPT decoder for gaze prediction.
\textbf{ViTGaze}~\cite{song2024vitgaze} uses a pretrained Vision Transformer and extracts human--scene interactions from self-attention maps using a 4D interaction encoder with 2D spatial guidance, followed by deconvolutional layers for gaze prediction.
For social gaze evaluation, these gaze models are used to derive social gaze predictions from either contextual features or predicted gaze locations, following the evaluation protocol of CoSI-Gaze~\cite{chang2026cosigaze}.

\subparagraph{Performance}

The performance was evaluated on an NVIDIA GeForce RTX 3090 with 23.69 GiB of GPU memory. Under this hardware configuration, the module requires an average of 30 ms per image for gaze following only, and 32.3 ms per image for joint gaze following and social gaze prediction. On the DyGaze test set, the gaze-following module achieves the best AUC of 0.95 and the best normalized $L_2$ distance of 0.12. For pair-wise social gaze prediction, the module achieves an average classification accuracy of 0.69, with F1-scores of 0.73, 0.67, 0.63, 0.62, and 0.74 for Share, Mutual, Single, Miss, and Void, respectively.

\begin{table}[htbp]
    \centering
    \caption{Gaze following results on DyGaze. Higher is better for AUC, while lower is better for L2 distance.}
    \label{tab:dygaze_results}
    \begin{tabular}{lcc}
        \toprule
        \textbf{Method} & \textbf{AUC $\uparrow$} & \textbf{L2 Dist. $\downarrow$} \\
        \midrule
        VideoAtt~\cite{chong2020videoatt}       & 0.94 & 0.14 \\
        Sharingan~\cite{tafasca2024sharingan}   & 0.93 & \textbf{0.12} \\
        ViTGaze~\cite{song2024vitgaze}          & 0.92 & 0.16 \\
        \textbf{CoSI-Gaze~\cite{chang2026cosigaze}} 
        & \textbf{0.95} 
        & \textbf{0.12} \\
        \bottomrule
    \end{tabular}
\end{table}

\begin{table}[htbp]
    \centering
    \caption{Class-wise social gaze prediction performance on DyGaze. Metrics include Precision (p.), Recall (r.) and F1 score (f1.). Best results are highlighted in \textbf{bold}.}
 \resizebox{\textwidth}{!}{
    \begin{tabular}{l|ccc|ccc|ccc|ccc|ccc|c}
          \hline
        \centering    
    \multirow{2}{*}{Method} 
    & \multicolumn{3}{c|}{Share} 
    & \multicolumn{3}{c|}{Mutual}
    & \multicolumn{3}{c|}{Single} 
    & \multicolumn{3}{c|}{Miss} 
    & \multicolumn{3}{c|}{Void}
    & \\
    & 
    p.& r. & f1. &
    p.& r. & f1. &
    p.& r. & f1. &
    p.& r. & f1. &
    p.& r. & f1. &
    Avg. Acc. \\
    \hline
    \multicolumn{17}{l}{Social Gaze Prediction obtained by: Context}\\
    \hline
    VideoAtt ~\cite{chong2020videoatt}
    & 0.47 & 0.15 & 0.23 & 0.64 & 0.74 & 0.69 & 0.43 & 0.55 & 0.48 & \textbf{0.62} & \textbf{0.62} & \textbf{0.62} & 0.53 & 0.67 & 0.59 & 0.55 \\
    
    UnifiedGaze ~\cite{gupta2024unified} 
    & 0.65 & 0.44 & 0.53 & \textbf{0.68} & 0.74 & 0.71 & 0.51 & 0.54 & 0.53 & 0.56 & 0.46 & 0.50 & 0.56 & 0.70 & 0.62 & 0.59 \\
    
    ViTGaze ~\cite{song2024vitgaze}
    & 0.71 & 0.66 & 0.69 & 0.63 & 0.85 & \textbf{0.72} & 0.62 & 0.59 & 0.60 & \textbf{0.62 }& 0.59 & 0.60 & \textbf{0.73} & 0.65 & 0.69 & 0.67 \\
    \hline
    \multicolumn{17}{l}{Social Gaze Prediction obtained by: Spatial}\\
    \hline
    VideoAtt ~\cite{chong2020videoatt}
    & 0.68 & 0.21 & 0.33 & 0.54 & 0.91 & 0.68 & 0.43 & 0.50 & 0.46 & 0.59 & 0.53 & 0.56 & 0.59 & 0.63 & 0.61 & 0.56 \\

    Sharingan ~\cite{tafasca2024sharingan}
    & 0.72 & 0.33 & 0.45 & 0.52 & 0.93 & 0.67 & 0.42 & 0.44 & 0.43 & 0.54 & 0.53 & 0.53 & 0.67 & 0.60 & 0.63 & 0.57 \\
    
    ViTGaze ~\cite{song2024vitgaze}
    & 0.75 & \textbf{0.67} & 0.71 & 0.45 & \textbf{0.97} & 0.62 & 0.53 & 0.38 & 0.44 & 0.53 & 0.38 & 0.44 & 0.66 & 0.47 & 0.55 & 0.57 \\
    \hline
    \multicolumn{17}{l}{Social Gaze Prediction obtained by: Both}\\
    \hline
    \textbf{CoSI-Gaze ~\cite{chang2026cosigaze}} 
    & \textbf{0.84} & 0.64 & \textbf{0.73} & 0.61 & 0.75 & 0.67 & \textbf{0.63} & \textbf{0.64} & \textbf{0.63} & \textbf{0.62} & \textbf{0.62} & \textbf{0.62} & \textbf{0.73} & \textbf{0.75} & \textbf{0.74} & \textbf{0.69} \\
     \hline
\end{tabular}
}
\label{tab:dyadic_gaze_details}
\end{table}

\section{Usage}
\label{sec:usage}
This section introduces the basic usage of SeetaPsych, covering its configuration mechanism, installation and deployment, Python APIs, and WebUI. It explains how algorithms and attributes are organized, how computation graphs are constructed, and how pipelines can be executed programmatically or through a visual interface.
The core SeetaPsych library is available at
\href{https://github.com/seetapsych/seetapsych-lib}{https://github.com/seetapsych/seetapsych-lib},
where readers can find further implementation details and project documentation.
Future updates and development documentation will also be released through the open-source project.

\subsection{Quick Start}

The project uses configuration files to describe the available algorithms and the attributes that each algorithm can produce.
An attribute represents the output of an algorithm or processing method.

\begin{figure}[htbp]
    \centering
    \includegraphics[width=0.9\linewidth]{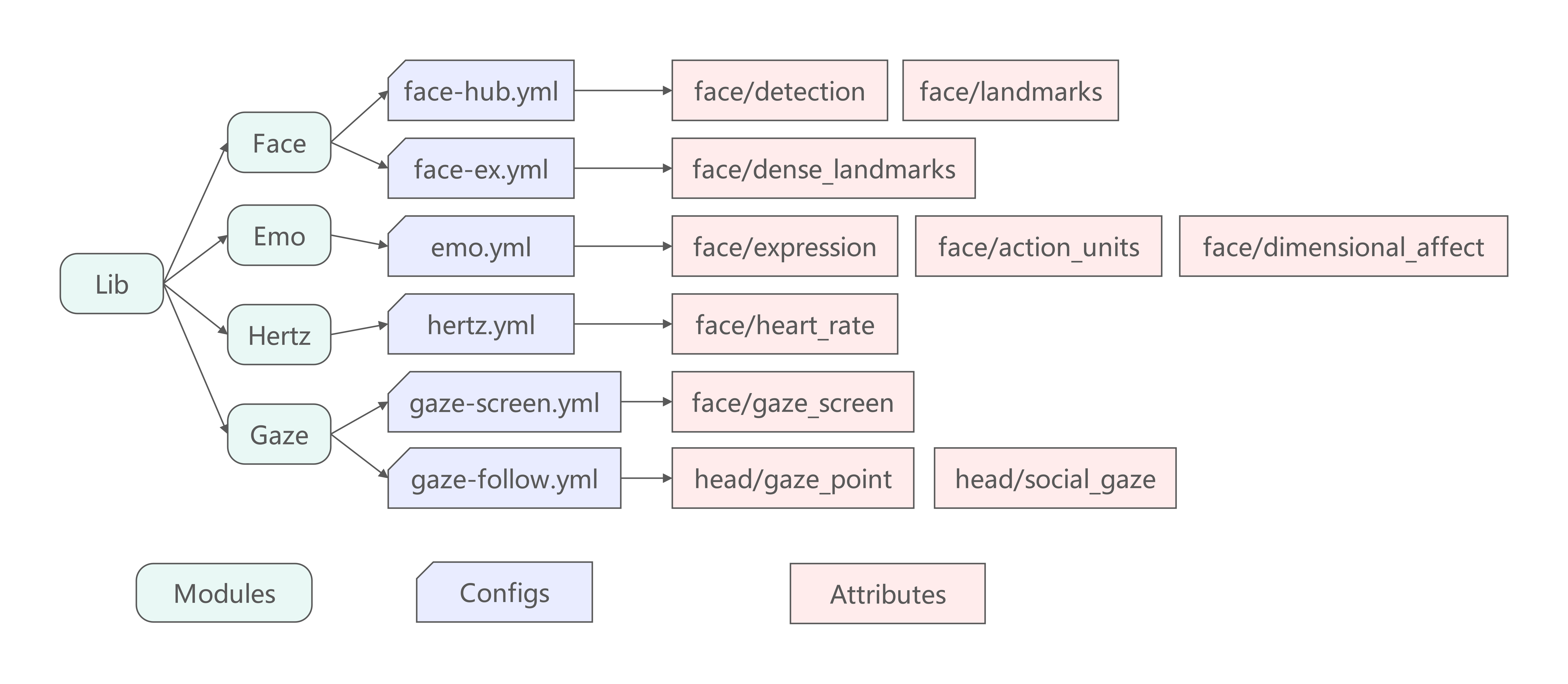}
    \caption{Configuration files(yml) and their corresponding attributes
    }
    \label{fig:algorithm_configuration}
\end{figure}

Figure~\ref{fig:algorithm_configuration} illustrates how the algorithms and attributes are described through configuration files(yml). For example, the algorithms and attributes about face preprocessing is describe in \texttt{face-hub.yml} and \texttt{face-ex.yml}. The face-hub consists of two algorithms face detection and facial landmarks detection.  They respectively produces \texttt{face\_detection} attribute and \texttt{face\_landmarks} attributes. The \texttt{face\_detection} attribute containing the detected face bounding box and its confidence score as below:
\begin{center}
\begin{minipage}{0.6\linewidth}
\begin{Verbatim}[
    frame=single,
    framesep=3mm,
    fontsize=\small
]
{
  "face_detection": [
    {
      "xyxy": [
        128.77244567871094,
        158.99908447265625,
        286.5462951660156,
        369.4015808105469
      ],
      "score": 0.8021643161773682
    }
  ]
}
\end{Verbatim}
\end{minipage}
\end{center}

These configuration files (yml) are part of the framework's internal management mechanism and normally do not need to be edited by users. The configuration files used by the framework are maintained in the \url{https://github.com/seetapsych/seetapsych-configs} repository and are automatically obtained when required.

Each attribute may depend on one or more algorithm modules for computation. Users only need to specify the attributes they want to obtain. Based on these requested attributes and their dependencies, the framework automatically selects the required modules and organizes them into a computation graph.

\begin{figure}[htbp]
    \centering
    \includegraphics[width=0.9\linewidth]{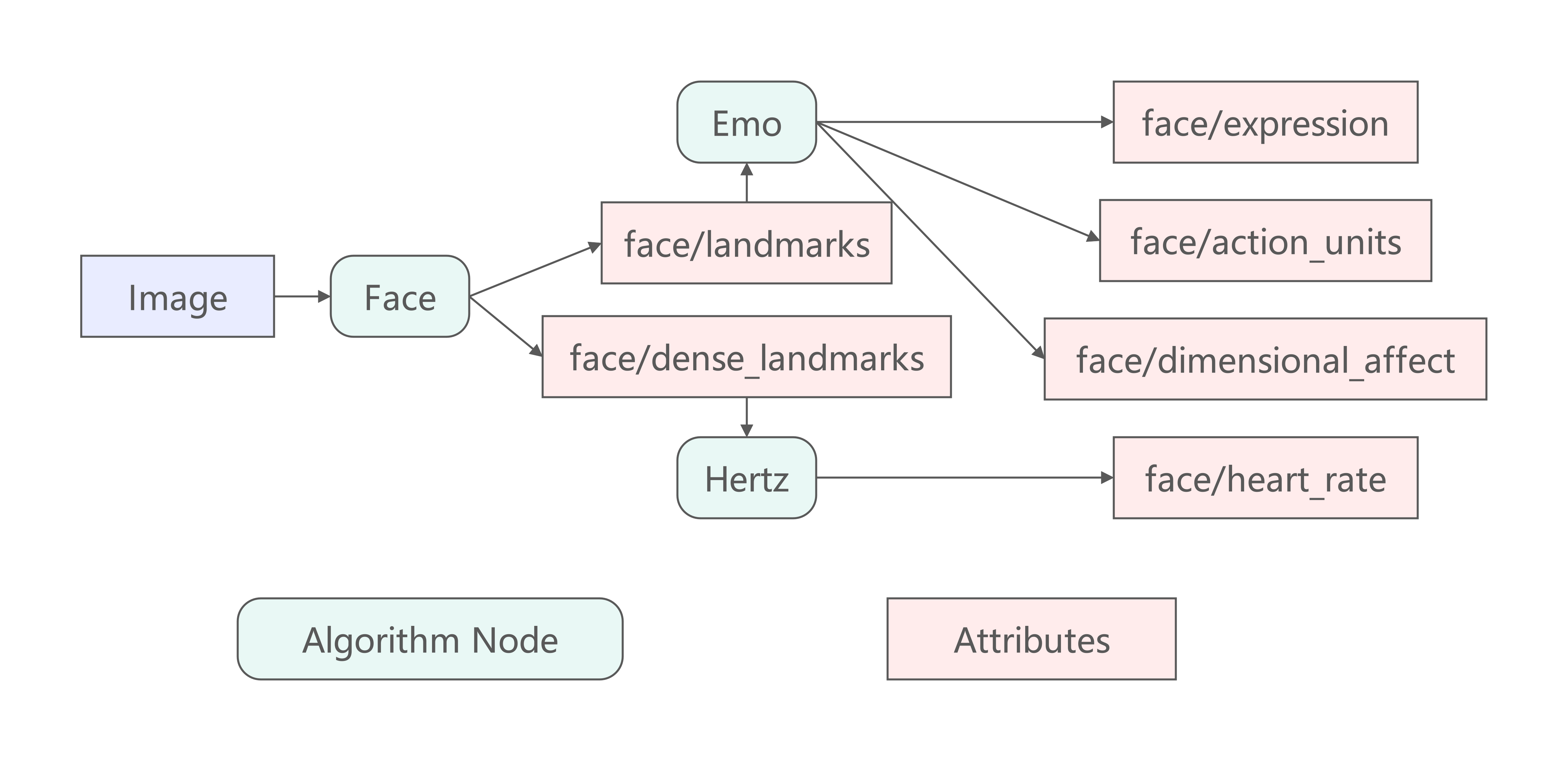}
    \caption{Example of a computation graph constructed from the requested attributes
    }
    \label{fig:computation_graph}
\end{figure}

The computation graph is executed by a \texttt{Runner}, which can be used to analyze images or videos and produce the requested attributes. By default, the \texttt{Runner} automatically detects the available hardware environment and prioritizes GPU acceleration for algorithm inference when a supported GPU is available.
Figure~\ref{fig:computation_graph} shows an example of a computation graph. 
In this example, the input image is first processed by the Face module to extract facial information, including sparse landmarks and dense landmarks. The Emo module uses sparse facial landmarks to produce expression, action unit, and dimensional affect attributes, while the Hertz module uses dense facial landmarks to estimate the heart-rate attribute.

\subsection{Installation and deployment}

Install the basic runtime environment:

\begin{center}
\begin{minipage}{0.9 \linewidth}
\begin{Verbatim}[
    frame=single,
    framesep=3mm,
    fontsize=\footnotesize
]
uv pip install 'seetapsych-lib[webui]' seetapsych-attributes seetapsych-configs
\end{Verbatim}
\end{minipage}
\end{center}

The basic runtime environment provides the core components required to run SeetaPsych. Algorithm-specific configuration files are maintained separately and need to be downloaded before the corresponding algorithms can be used.

The configuration files for each algorithms are available from \url{https://github.com/seetapsych/seetapsych-configs}. To download the latest configuration files for all available algorithms, run:

\begin{center}
\begin{minipage}{0.4\linewidth}
\begin{Verbatim}[
    frame=single,
    framesep=3mm,
    fontsize=\footnotesize
]
seetapsych-manager download
\end{Verbatim}
\end{minipage}
\end{center}

The \texttt{seetapsych-manager} command-line tool and \texttt{seetapsych-webui} are installed automatically with \texttt{seetapsych-lib[webui]}.
By default, this command downloads the latest configuration files to the framework's default configuration directory.

After the download is complete, run the following command to inspect the available algorithm configurations:

\begin{center}
\begin{minipage}{0.4\linewidth}
\begin{Verbatim}[
    frame=single,
    framesep=3mm,
    fontsize=\footnotesize
]
seetapsych-manager show
\end{Verbatim}
\end{minipage}
\end{center}

\subsection{APIs}

The Python API can be used to construct and execute analysis pipelines for different attributes.

The  example Code~\ref{lst:seetapsych_face_detection} below demonstrates how to create a pipeline for the \texttt{face/detection} attribute, resolve its dependencies, prepare the required runtime resources, and execute the pipeline on an image.
The \texttt{Factory} manages the available algorithm modules. A \texttt{Pipeline} is then created by specifying the attributes to be computed. The pipeline can automatically resolve attribute dependencies, install missing requirements, and download the required models.

After the pipeline is prepared, a \texttt{Runner} is used to execute it. \texttt{ParallelRunner} can also be used when parallel execution is required.
Available attributes provided by the installed algorithm modules can be inspected using the \texttt{seetapsych-manager show} command. The output structure and field definitions for each attribute are documented in the \texttt{seetapsych-attributes} repository:
\url{https://github.com/seetapsych/seetapsych-attributes}.

\begin{lstlisting}[
    language=Python,
    caption={Example of constructing and executing a face-detection pipeline with SeetaPsych.},
    label={lst:seetapsych_face_detection}
]
# -*- coding: utf-8 -*-

import json
import cv2

from seetapsych_lib.runtime.factory import Factory
from seetapsych_lib.runtime.pipeline import Pipeline
from seetapsych_lib.runtime.runner import Runner
from seetapsych_lib.runtime.parallel_runner import ParallelRunner


def main():
    # All installed algorithm modules are loaded by default during initialization.
    # You can use the `load_xxx_module(s)` methods to load specific algorithm modules.
    factory = Factory()

    # Quickly build a workflow and declare the attribute to compute
    # as the face feature 'face/detection'.
    # You can view all available attributes of installed algorithms
    # using the `seetapsych-manager show` command.
    # Result fields for attributes can be found at:
    # https://github.com/seetapsych/seetapsych-attributes
    pipeline = Pipeline(factory, attributes=['face/detection'])

    # Check for dependencies or missing issues that need to be resolved with solve().
    print(pipeline.problem())

    # Resolve workflow dependencies, automatically add face detection
    # and corresponding models.
    pipeline.solve()

    # Check for runtime environment issues that require installation
    # or download to fix.
    print(pipeline.satisfied())

    # Install missing dependencies required for the current pipeline to run.
    pipeline.install_requirements()

    # Download missing models required for the pipeline to run.
    pipeline.cache_models()

    # Create a basic executor.
    runner = Runner(pipeline)

    # Or create a parallel executor.
    # runner = ParallelRunner(pipeline)

    # Run the algorithm.
    report = runner.run(data={
        'default': cv2.imread('image.jpg')
    })

    # Print the execution results.
    print(json.dumps(report, indent=2, ensure_ascii=False))


if __name__ == '__main__':
    main()
\end{lstlisting}

\subsection{WebUI}

A WebUI is provided for quickly configuring and executing analysis pipelines. Start the WebUI with the following command:
\begin{center}
\begin{minipage}{0.3\linewidth}
\begin{Verbatim}[
    frame=single,
    framesep=3mm,
    fontsize=\footnotesize
]
seetapsych-webui
\end{Verbatim}
\end{minipage}
\end{center}

Additional algorithm configuration files can be specified when starting the WebUI. For example, if you need to specify additional algorithm description files, you can add the specified directory parameter by command:
\begin{center}
\begin{minipage}{0.5\linewidth}
\begin{Verbatim}[
    frame=single,
    framesep=3mm,
    fontsize=\footnotesize
]
seetapsych-webui --dirs <dir>
\end{Verbatim}
\end{minipage}
\end{center}

Configuration files can be specified in several ways:

\begin{itemize}
    \setlength{\itemsep}{0pt}
    \setlength{\parskip}{0pt}
    \item \texttt{--dirs}: Specify directories containing configuration files.
    \item \texttt{--files}: Specify individual configuration files.
    \item \texttt{--urls}: Specify URLs of configuration files.
\end{itemize}

\begin{figure}[htbp]
    \centering
    \includegraphics[width=0.9\linewidth]{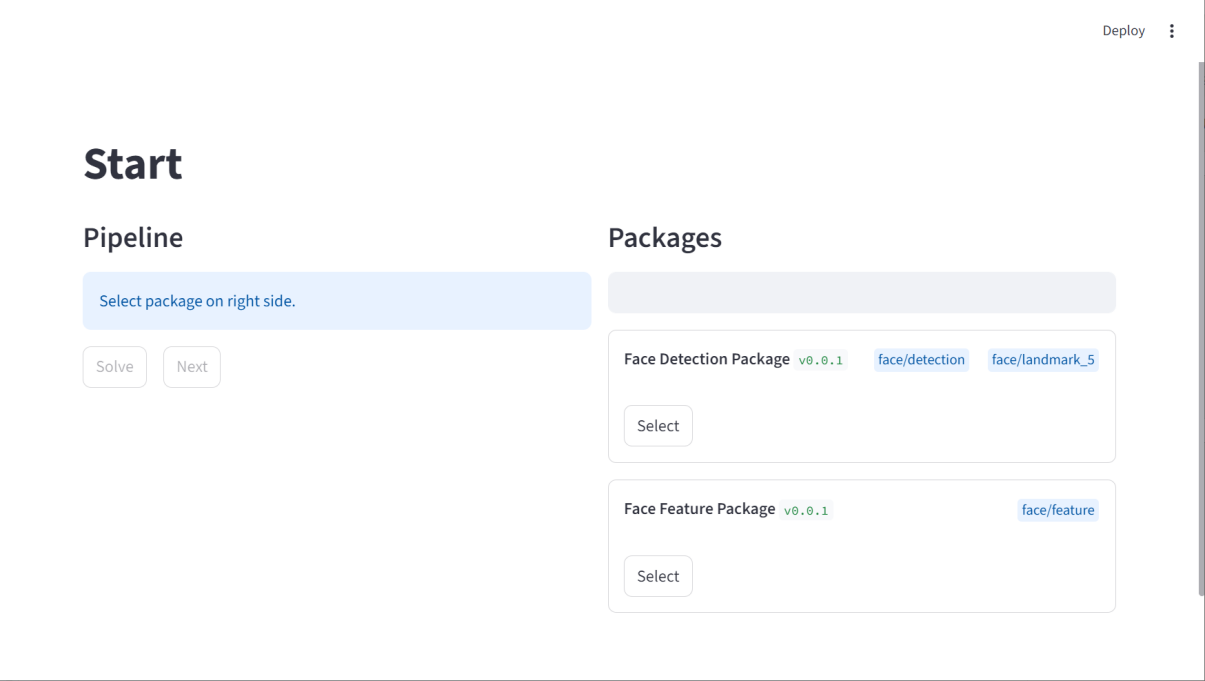}
    \caption{The opened WebUI interface}
    \label{fig:webui_interface}
\end{figure}

After startup, the WebUI opens automatically in the default browser. 
Or you can manually open the browser and enter the URL:\href{http://localhost:8501}{http://localhost:8501}.
The opened web page is shown as Figure~\ref{fig:webui_interface}.

\begin{figure}[htbp]
    \centering
    \includegraphics[width=0.9\linewidth]{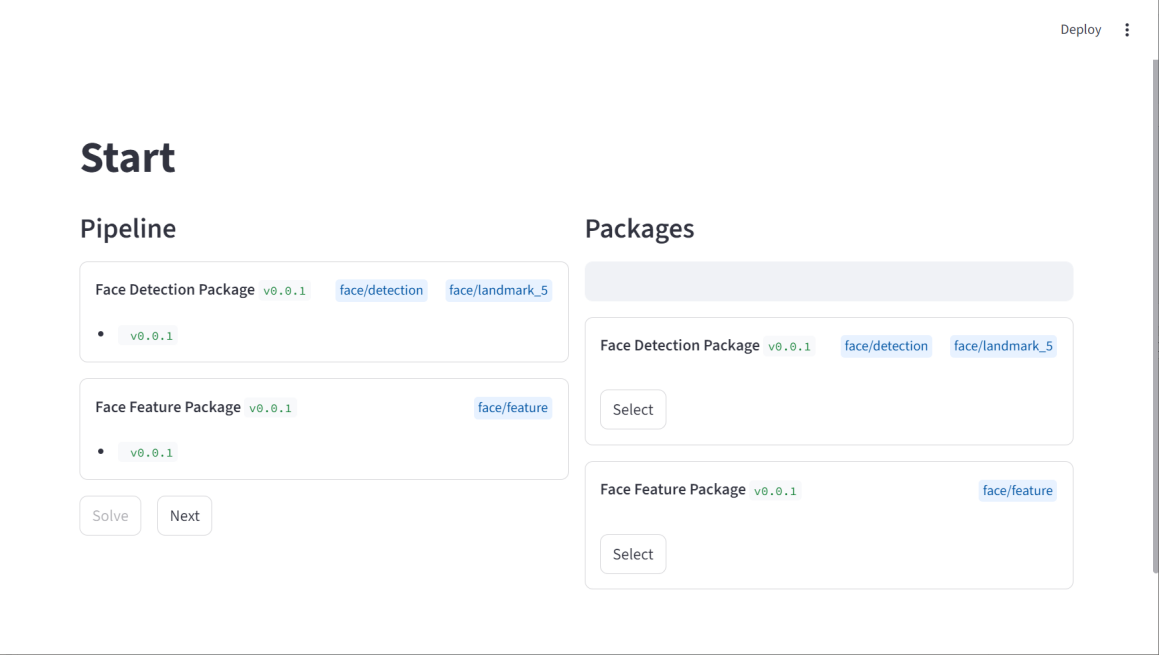}
    \caption{An example of visual pipeline construction.}
    \label{fig:visual_pipeline_construction}
\end{figure}

The WebUI provides a visual interface for constructing a Pipeline. You can select the attributes to be computed, choose the corresponding algorithm modules and models, and configure the required parameters. Based on these selections, the WebUI builds the Pipeline for subsequent execution. Figure~\ref{fig:visual_pipeline_construction} shows an example.

\begin{figure}[htbp]
    \centering
    \includegraphics[width=0.9\linewidth]{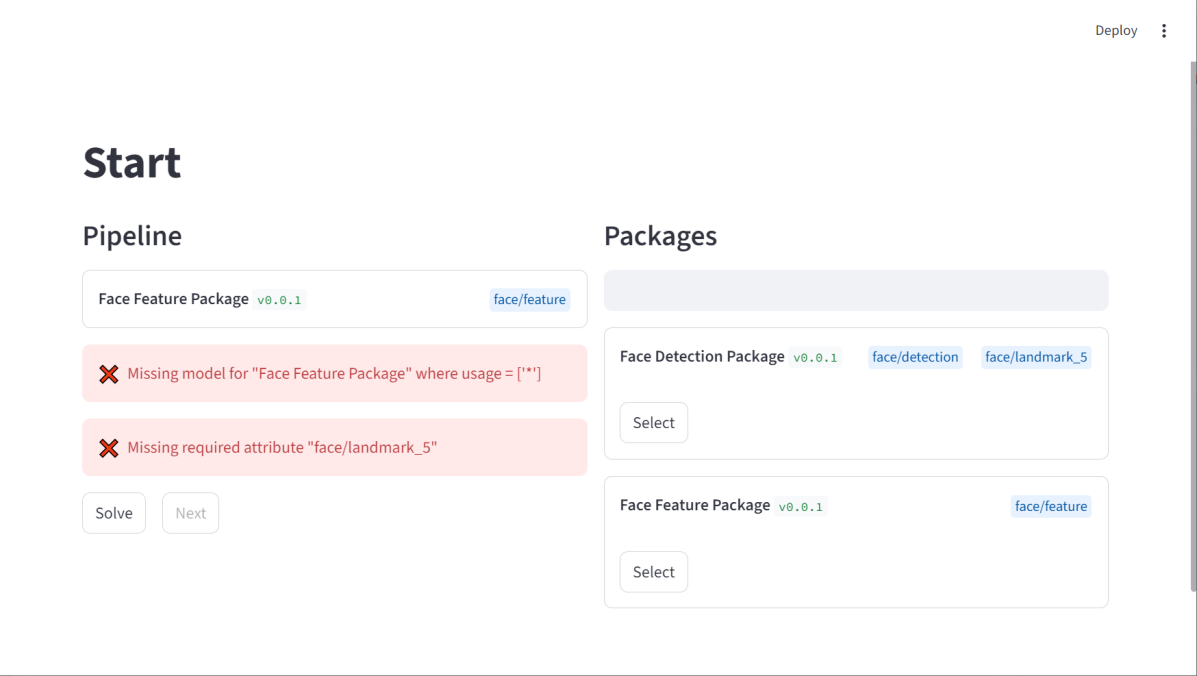}
    \caption{An example of automatic missing dependency resolution.}
    \label{fig:automatic_dependency_resolution}
\end{figure}

During pipeline construction, most unresolved dependencies can be detected and resolved automatically. Clicking the \texttt{Solve} button analyzes the current pipeline and completes the required dependency relationships. Figure~\ref{fig:automatic_dependency_resolution} shows an example.

\begin{figure}[htbp]
    \centering
    \includegraphics[width=0.9\linewidth]{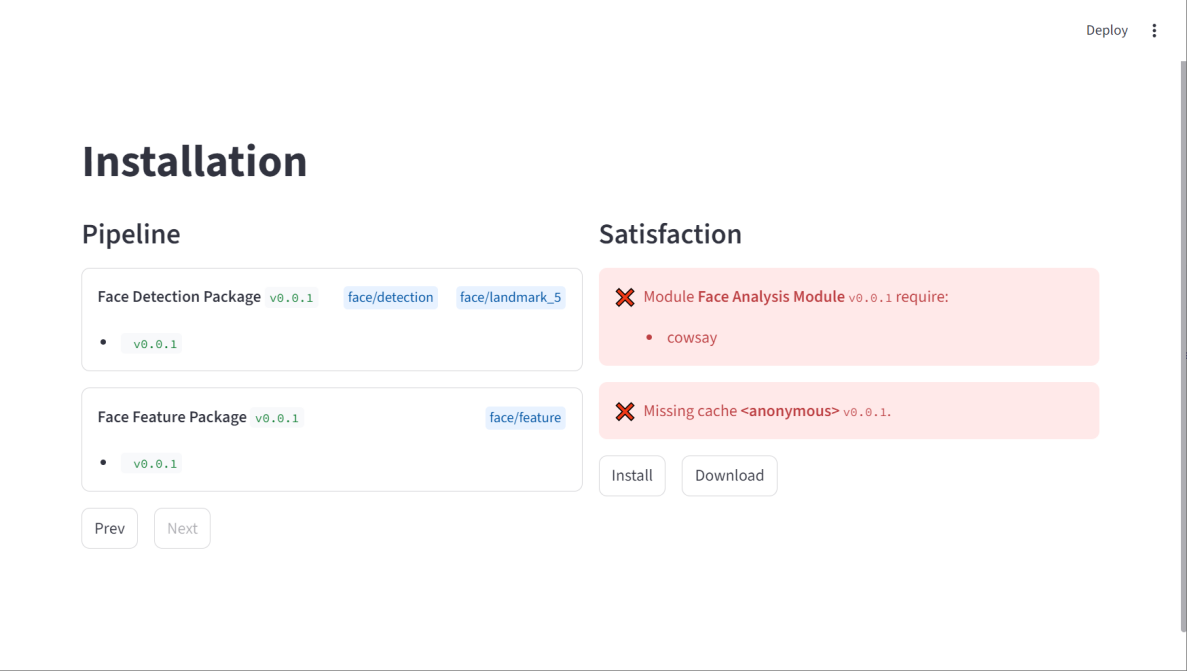}
    \caption{An example of automatic download of dependency libraries and models.}
    \label{fig:automatic_dependency_download}
\end{figure}

Required libraries and model files can also be installed or downloaded directly through the WebUI, simplifying the preparation of the runtime environment. Figure~\ref{fig:automatic_dependency_download} shows an example.

\begin{figure}[htbp]
    \centering
    \includegraphics[width=0.9\linewidth]{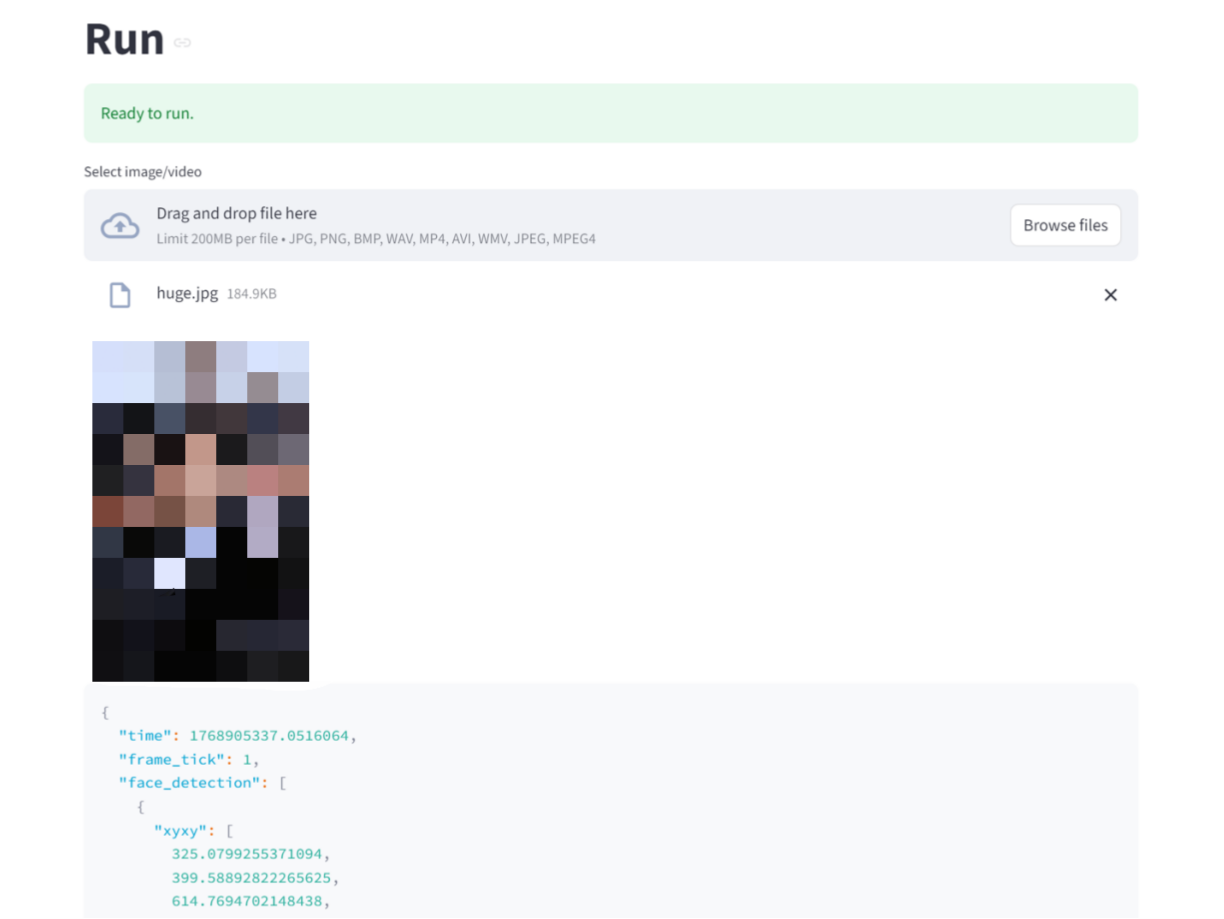}
    \caption{An example of uploading an image for inference.}
    \label{fig:upload_media_inference}
\end{figure}

After configuration, the Pipeline can be saved to a file and loaded again for later use, avoiding the need to reconstruct the same workflow.
Images or videos can then be uploaded to execute the configured Pipeline. The analysis results are returned in JSON format and can also be exported as CSV files for subsequent analysis. Figure~\ref{fig:upload_media_inference} shows an example of uploading an image for inference.

\section{Conclusion}

SeetaPsych v1.0 provides a unified and extensible toolkit for behavior-based psychological measurement from visual data, integrating four major functions together with supporting human-centric preprocessing tools. Its modular Pipeline/Runner framework, Python APIs, and interactive WebUI support flexible programmatic and interactive analysis workflows.

Although SeetaPsych v1.0 supports several behavior-based physiological measurements, further improvements are needed to support reliable use across more diverse populations, acquisition conditions, and experimental settings. For face-based emotion analysis, an important direction is to improve recognition accuracy and cross-domain generalization while retaining the current multi-task capability for expression recognition, action unit detection, and valence-arousal estimation, together with practical inference efficiency. For screen point-of-gaze estimation, future work will aim to improve robustness across wider ranges of head pose and viewing distance, expand the spatial coverage of reliable gaze estimation, and introduce personalized calibration and adaptation to reduce subject-specific errors associated with factors such as eyeglasses and individual facial or ocular characteristics. Heart-rate estimation also remains sensitive to motion-related disturbances, particularly those caused by head pose changes and facial motion. More broadly, the included algorithms will benefit from continued evaluation and refinement across diverse populations, devices, environments, and less constrained recording conditions.

SeetaPsych will continue to evolve at both the toolkit and framework levels. Potential directions include extending the range of measurable behavioral signals, particularly body-related cues such as subtle body movements and other nonverbal behaviors, and incorporating additional algorithms and models. The WebUI and management tools will be further improved to support a broader range of input formats and interaction modes, making them better suited to practical experimental and analysis scenarios while simplifying pipeline construction, dependency and model management, and result inspection. Another possible direction is to provide machine-readable skills and related interfaces that facilitate the discovery, configuration, and invocation of SeetaPsych capabilities by AI agents and agent-based tools.

Finally, the analysis of human appearance, behavior, attention, emotion-related cues, and physiological signals raises important ethical and privacy considerations. Responsible use of SeetaPsych requires appropriate data collection procedures, informed consent, data protection, and careful interpretation of model outputs. Researchers and practitioners should also ensure compliance with applicable ethical guidelines and privacy regulations when using the toolkit in research or real-world applications.
\clearpage
\bibliographystyle{unsrt}
\bibliography{references}

\end{document}